\documentclass[journal]{IEEEtran}
\usepackage{float}
\usepackage[symbol]{footmisc}
\usepackage{lineno,hyperref}
\usepackage{epstopdf}
\usepackage{multirow}
\usepackage{siunitx}
\usepackage{booktabs}
\usepackage{comment}
\usepackage{times}
\usepackage{epsfig}
\usepackage{graphicx}
\usepackage{amsmath}
\usepackage{amssymb}
\usepackage{multirow}
\usepackage{float}
\usepackage{xcolor}
\usepackage{graphicx}
\usepackage{graphics}
\usepackage{subcaption}
\usepackage{bigstrut}
\usepackage{makecell}
\usepackage{booktabs}
\usepackage{bibentry}
\usepackage{epstopdf}
\usepackage{tabularx}
\usepackage{balance}
\usepackage{comment}
\usepackage{esvect} 
\usepackage{algorithm}
\usepackage{algorithmic}
\usepackage{pifont}
\usepackage{paralist}
\usepackage{url}
\usepackage{placeins}

\newcommand{\RED}[1]{\textcolor{black}{#1}}

\newcolumntype{C}[1]{>{\centering\arraybackslash}p{#1}}

\def \R {\mathbb{R}}

\def \N {\mathcal{N}}

 \makeatletter
    \newcommand{\thickhline}{%
        \noalign {\ifnum 0=`}\fi \hrule height 1pt
        \futurelet \reserved@a \@xhline
    }
    \newcolumntype{"}{@{\vrule width 1pt}}
    \makeatother

\ifCLASSINFOpdf
\else
\fi

\begin{document}
\title{Graph Convolution Based Efficient Re-Ranking for Visual Retrieval}
\author{Yuqi~Zhang$^{*}$,Qi~Qian$^{*}$, Hongsong~Wang$^{* \dagger}$, Chong~Liu, Weihua~Chen, Fan~Wang
\thanks{${\dagger}$~Corresponding author. *~indicates equal contributions.}
\thanks{This research is supported in part by Southeast University Start-Up Grant for New Faculty under Grant RF1028623063. }
\thanks{Y.~Zhang, Q.~Qian, W.~Chen, and F.~Wang are with Machine intelligence Technology Lab, Alibaba Group (email: qi.qian@alibaba-inc.com;kugang.cwh@alibaba-inc.com).}
\thanks{H.~Wang is with Department of Computer Science and Engineering, Southeast University, Nanjing, China (email: hongsongwang@seu.edu.cn).}
}
\maketitle

\begin{abstract}
Visual retrieval tasks such as image retrieval and person re-identification (Re-ID) aim at effectively and thoroughly searching images with similar content or the same identity. 
After obtaining retrieved examples, re-ranking is a widely adopted post-processing step to reorder and improve the initial retrieval results by making use of the contextual information from semantically neighboring samples. 
Prevailing re-ranking approaches update distance metrics and mostly rely on inefficient crosscheck set comparison operations while computing expanded neighbors based distances. In this work, we present an efficient re-ranking method which refines initial retrieval results by updating features. Specifically, we reformulate re-ranking based on Graph Convolution Networks (GCN) and propose a novel Graph Convolution based Re-ranking (GCR) for visual retrieval tasks via feature propagation. To accelerate computation for large-scale retrieval, a decentralized and synchronous feature propagation algorithm which supports parallel or distributed computing is introduced. In particular, the plain GCR is extended for cross-camera retrieval and an improved feature propagation formulation is presented to leverage affinity relationships across different cameras. It is also extended for video-based retrieval, and Graph Convolution based Re-ranking for Video (GCRV) is proposed by mathematically deriving a novel profile vector generation method for the tracklet. 
Without bells and whistles, the proposed approaches achieve state-of-the-art performances on seven benchmark datasets from three different tasks, i.e., image retrieval, person Re-ID and video-based person Re-ID.
\RED{Code is publicly available: \textcolor{red}{\url{github.com/WesleyZhang1991/GCN_rerank}}.}
\end{abstract}

\begin{IEEEkeywords}
Visual retrieval, re-ranking, person re-identification, video-based person re-identification
\end{IEEEkeywords}

%
\IEEEpeerreviewmaketitle

\section{Introduction}\label{sec:introduction}
\IEEEPARstart{G}{iven} a query image, the goal of visual retrieval is to effectively and efficiently find relevant images that share similar content or depict the same instance from a large-scale visual corpus. The typical tasks include content-based image retrieval, person re-identification (Re-ID) and vehicle Re-ID. Visual retrieval has become an increasingly important field of research with the fast development of Internet and the universal popularity of digital cameras. 
Although this problem has been extensively explored by researchers over the past decades, it is still very challenging to trade between the efficiency and accuracy while deploying a retrieval method in realistic applications. 

When training models for image retrieval or person Re-ID, most approaches only consider the identity label of an image and the relationship between two images while ignoring the contextual information among multiple relevant images. Re-ranking approaches~\cite{mei2014multimedia,tao2011visual} mitigate this issue by utilizing global similarity measures which could analyze an ample amount of relevant images and take the underlying global manifold structure into account. The re-ranking process reorders the retrieved results based on the information or specific patterns mined from the initial ranked list. With additional computation cost, re-ranking as a post-process step has been experimentally validated to be very effective for content-based image retrieval, person or vehicle re-identification and so on.

\begin{figure} 
	\centering
	\includegraphics[width=0.5\textwidth]{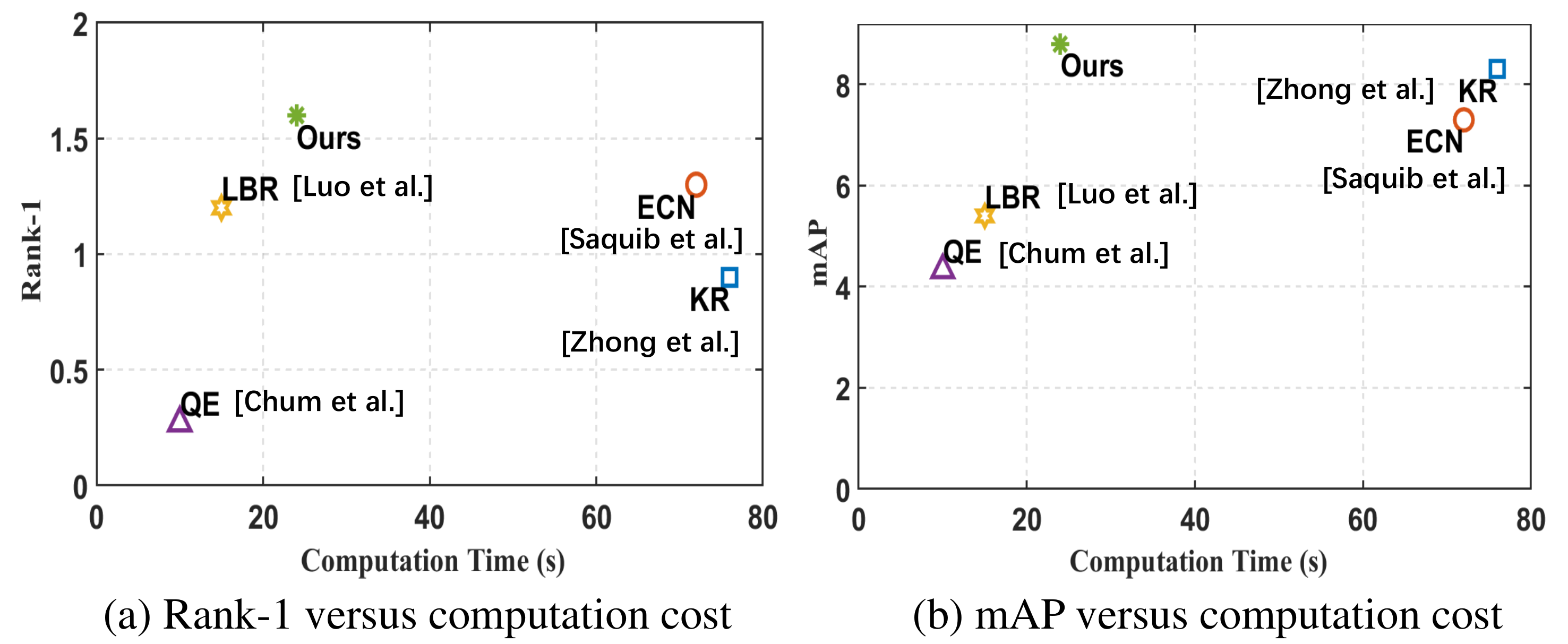}
	\caption{Comparison of the computation time and the gained performance of different re-ranking methods for person Re-ID on the Market-1501 dataset. The computation time of our approach is only one third of that of KR~\cite{zhong2017re} or ECN~\cite{saquib2018pose}, while the performance is even dramatically improved. Although some methods such as QE~\cite{chum2007total} enjoy less computation time, their retrieval performances are rather limited to meet the realistic demand.}
	\label{fig:acc_time}
\end{figure}
\begin{figure*} 
	\centering
	\includegraphics[width=1.0\textwidth]{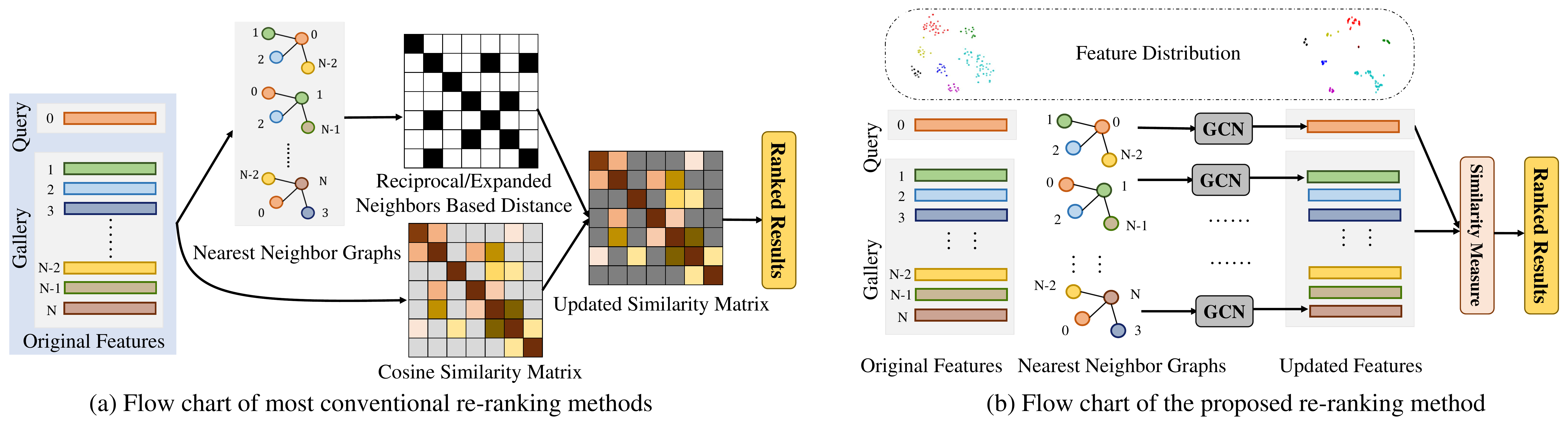}
	\caption{A comparison between most conventional re-ranking methods and the proposed GCR. Previous methods such as KR~\cite{zhong2017re} and ECN~\cite{saquib2018pose} directly calculate the similarity matrix based on nearest neighbors and their reciprocal or expanded neighbors. In contrast, our approach indirectly refine initial ranking results by updating and improving the features.}
	\label{fig:ideas}
\end{figure*}
Previous re-ranking methods can be roughly categorized into two types: learning based and non-learning based approaches. The former one jointly optimizes image features and pairwise relationships in a single network~\cite{shen2018deep,shen2018person,wu2020adaptive}.
These approaches are tightly restricted to the feature extraction backbone, and cannot be used as an off-the-shelf re-ranking module. The latter one only focuses on the post-processing step and is more flexible for different applications. Typical non-learning based approaches are Query Expansion (QE)~\cite{chum2007total}, K-Reciprocal Encoding (KR)~\cite{zhong2017re} and ECN~\cite{saquib2018pose}. Although KR and ECN evidently improve the retrieval performance, they are inefficient for large-scale applications due to the requirement of computing sophisticated distance metrics between the query and all samples in the database with respect to reciprocal or expanded neighbors. 

Graph Convolution Networks (GCN) have been applied for content-based image retrieval and person Re-ID~\cite{shen2018deep, shen2018person,wu2020adaptive}, mostly in order to learn pairwise similarity information or contextual interactions. 
These approaches appropriately utilize relationships between probe and gallery images to learn powerful features.
However, existing GCN based re-ranking methods are not plug-and-play as they often combine the GCN based contextual relationship with the CNN based appearance feature learning to form a unified framework. 
Recently, the standalone re-ranking process in image retrieval is also re-formulated from a GCN perspective~\cite{zhang2020understanding}. Although this approach shares some similarities with ours in the target of GCN based re-ranking, there exist considerable differences in mathematical modelling of relation graph construction and feature propagation formulas as well as potential applications.


This paper targets at a more scalable and efficient re-ranking that is flexible for various visual retrieval tasks. We novelly establish the intrinsic connection between GCN and nearest neighbors based re-ranking, and propose a new re-ranking paradigm by updating features based on GCN. 
As shown in Fig.~\ref{fig:acc_time}, the proposed re-ranking could improve the performance over the widely used KR~\cite{zhong2017re} and ECN~\cite{saquib2018pose} while being much more efficient. 
Different from previous methods that handle the similarities directly, our approach indirectly refines the initial search results by updating and improving the features. A brief comparison between our approach and conventional re-ranking methods is illustrated in Fig.~\ref{fig:ideas}. 
Different from conventional methods that refine the similarity matrix, our approach improves the features by aligning them across similar instances.
Most conventional re-ranking methods require the storage of the intermediate similarity matrices. When the size of the gallery set is very large, these similarity matrices become extremely large and bring heavy memory burdens. In contrast, our approach only stores the feature matrices which are much smaller and more memory friendly. 
Additionally, previous approaches inevitably rely on inefficient set comparison operations which are difficult to be accelerated. Our approach only depends on matrix operations which can be easily deployed on the GPU.

More specifically, we propose a novel Graph Convolution based Re-ranking (GCR) for visual retrieval based on GCN.
In this framework, undirected graphs based on nearest neighbors are constructed to facilitate semantically contextual feature propagation.
To meet the demand for efficiency for large-scale retrieval tasks, a decentralized and synchronous feature propagation algorithm is presented to fit for parallel or distributed computing. 
Moreover, the proposed GCR is extended for two specific tasks: cross-camera retrieval and video-based retrieval. 
For the former task, an improved GCR is established by deliberately constructing cross-camera nearest neighbor graphs and aligning features from different cameras. For the latter task, a novel profile vector generation method is proposed to learn the profile vector for each tracklet, and the Graph Convolution based Re-ranking for Video (GCRV) is presented to fully utilize spatial and temporal contextual relationships.
The work is an extension of the published conference paper~\cite{zhang2022graph}. Compared to the preliminary version, the major differences are as follows.
\begin{compactitem}
	\item Compared with the previous GCR which is solely designed for person Re-ID, we present a more general re-ranking method for different visual retrieval tasks.  
	\item We propose a decentralized and synchronous feature propagation suitable for parallel or distributed computing.
	\item We derive the formulation of the profile vector generation from the perspective of feature alignment. 
	\item We conduct extensive experiments on four more benchmarks of visual retrieval, and provide more thorough analyses about efficiencies and parameter sensitivities.
\end{compactitem}

The rest of this paper is organized as follows. Section~\ref{sec:related} reviews the related work. Section~\ref{sec:method} elaborates feature propagation and the proposed re-ranking methods for different tasks. Section~\ref{sec:exp} provides extensive experimental results and analyses on different benchmark datasets. Section~\ref{sec:conclude} concludes this work with the future direction.

\section{Related Work}
\label{sec:related}
We briefly review approaches that are mostly related to ours from three aspects: person Re-ID, re-ranking for person Re-ID, and re-ranking for image retrieval. 

\subsection{Person Re-Identification}
Many recent works of person Re-ID focus on global or local feature learning.
The global approaches~\cite{sun2017svdnet, zheng2017discriminatively, chen2017multi, luo2019bag, luo2019strong} straightforwardly learn representations for the entire image. In contrast, local approaches~\cite{sun2018beyond, suh2018part, li2017learning, zhang2017alignedreid, zhang2019cross} learn representations for the image parts based on the assumption that person images are roughly aligned. Compared with global feature learning, local feature learning could alleviate the impact of occlusion or inaccurate detection.

The feature learning methods often use the Softmax loss or its modifications~\cite{liu2017sphereface, wang2018cosface, deng2019arcface}. 
Metric learning approaches such as the contrastive loss~\cite{hadsell2006dimensionality} and triplet loss~\cite{schroff2015facenet} are also applied for person Re-ID.
The widely used open-source re-ID strong baseline~\cite{luo2019bag} combines both softmax loss and triplet loss. Equipped with some tricks including random erasing~\cite{zhong2020random}, dropout~\cite{srivastava2014dropout} and so on, this baseline achieves competitive results compared with recent state-of-the-art approaches. A semi-supervised feature learning method is proposed to learn apparel-invariant feature embedding against appearance changes~\cite{yu2021apparel}. 
In our experiments of person Re-ID, this baseline model is used for feature extraction in order to make a fair comparison with other re-ranking methods.

Recently, there are also graphical models or Graph Convolution Networks (GCN) based feature learning approaches for person Re-ID~\cite{shen2018deep, shen2018person, wu2020adaptive}. Shen et al.~\cite{shen2018deep} apply similarity transformation on the graph to model the similarity relationship among the query image and the gallery images. 
Shen et al.~\cite{shen2018person} propose Similarity Guided Graph Neural Neural (SGGNN) to incorporate the pair-wise similarity information into the training process of person Re-ID. 
Wu et al.~\cite{wu2020adaptive} use GCN to learn the contextual interactions between the relevant regional features for video based person Re-ID.
However, these methods are often embedded into the training phase and thus could not serve as a re-ranking method in the post-process stage. Moreover, they do not take the cross-camera relationship into account. In this work, we reformulate the re-ranking based on GCN and also incorporate the cross-camera semantic similarities into the proposed re-ranking framework.

\subsection{Re-Ranking for Person Re-Identification}
Re-ranking is often used as a post-processing step to improve initial ranking results. The re-ranking methods can be generally split into two types: learning based and non-learning based approaches.

Learning based methods jointly optimize image features and pairwise relationships in a single network. Typical approaches are ~\cite{shen2018deep,shen2018person,luo2019spectral,wu2020adaptive}. 
For example, Luo et al.~\cite{luo2019spectral} propose a lightweight re-ranking method to make the underlying clustering structure around the query set more compact. 
However, most of these approaches are based on supervised similarity propagation which is integrated into the training pipeline. 
Conversely, our approach is a pure feature similarity propagation based and plug-and-play re-ranking technique, which is applicable for various visual retrieval tasks.

Non-learning based approaches only focus on the re-ranking post-processing. Since the original top-ranked data might be polluted by false positives, k-reciprocal nearest neighbors~\cite{jegou2007contextual, qin2011hello} can be regarded as highly relevant candidates. Zhong et al.~\cite{zhong2017re} introduce the concept of k-reciprocal nearest neighbor and design a re-ranking method for person Re-ID by combining the Jaccard distance of k-reciprocal encodings with the original feature distance. Sarfraz et al.~\cite{saquib2018pose} present a relatively effective re-ranking method which reinforces the original pairwise distance by aggregating distances between expanded neighbors of image pairs. 
To address the high computational complexity and unaffordable time cost for large-scale applications, Zhang et al.~\cite{zhang2020understanding} re-formulate the re-ranking process as a GCN function.
These approaches require sophisticated and complicated distance metrics rather than Euclidean distance with respect to reciprocal or expanded neighbors. In addition, most of these approaches could hardly achieve fast retrieval for massive data in real applications due to their high computational complexities. In contrast, our approach only relies on Euclidean distance for similarity measure which is more flexible and scalable to downstream tasks, and is specially designed for large-scale applications. 

\subsection{Re-Ranking for Image Retrieval}
Most approaches for image-retrieval often follow the retrieve-and-rerank paradigm which deploys a two-stage framework, i.e., first use global features to get the initial rank list, and then use local features to re-rank the candidates. 
Geometric verification is a popular image re-ranking approach and is widely used in both traditional work~\cite{philbin2007object} and recent deep learning based works~\cite{simeoni2019local,cao2020unifying}.
The DELG~\cite{cao2020unifying} jointly learns global and local features into a single deep model. However, the matching algorithm is still classic RANSAC which limits the performance and efficiency. SuperGlue~\cite{sarlin2020superglue} improves the local matching algorithm by solving a differentiable optimal transport problem with graph neuron networks. Reranking Transformer (RRT)~\cite{tan2021instance} uses transformers to learn matched or unmatched labels based on global and local features from two images. These approaches only consider pairwise local relationships and ignore informative relationships among multiple images. 
In addition, they are mostly developed for retrievals of rigid objects such as buildings and landmarks, and may not be applicable for other visual retrieval tasks such as person Re-ID.
This work is targeted at a general and effective re-ranking approach for different visual retrieval problems.

\subsection{\RED{Graph Structure Learning and Subspace Learning}} %
\RED{Graph Structure Learning (GSL) aims to jointly learn a graph structure and corresponding graph representations. A general pipeline of GSL has four steps: graph construction, structure modeling, message propagation and joint optimization learning~\cite{zhu2021survey}. For example, Pro-GNN~\cite{jin2020graph} jointly learns a structural graph and a robust GCN model from the perturbed graph.
The $k$-nearest neighbors and the GCN are usually used in graph construction and message propagation, respectively. For this reason, the proposed GCR can be regarded as a special type of the GSL where the joint optimization learning is omitted and the GCN parameters are fixed. }

\RED{	
The proposed GCN based re-ranking is also related to Subspace Representation Learning (SRL) which assumes a data instance can be expressed as a linear combination of other data instances from the same subspace. SRL maps high-dimensional features onto low-dimensional subspaces, and GCN can also be used to learn latent subspace presentations. Ji et al.~\cite{ji2017deep} introduce a novel self-expressive layer for unsupervised subspace clustering and directly learn the affinities between all data points. Zhang et al.~\cite{zhang2019neural} propose a collaborative learning paradigm for unsupervised subspace
clustering. A deep $k$-subspace clustering algorithm is presented to scale subspace clustering to
large-scale datasets~\cite{zhang2019scalable}.
}

\section{Proposed Approach}
\label{sec:method}
We first describe the proposed Graph Convolution based Re-ranking (GCR) from the perspective of Graph Convolution Network (GCN). 
Then we detail different feature propagation algorithms based on nearest neighbor graphs. 
Finally, we extend the basic approach for both cross-camera retrieval and video-based retrieval.

\subsection{Graph Convolution Based Re-Ranking} \label{plain_gcr}
The graph convolution operator in a standard Graph Convolution Networks (GCN)~\cite{kipf2016semi} layer can be written as
\begin{eqnarray}
	\mathbf{\widetilde{X}}=\mathbf{D}^{-\frac{1}{2}}\mathbf{AD}^{-\frac{1}{2}}\mathbf{XW}
\end{eqnarray}
where $\mathbf{X} \in \R^{n \times d}$ is the input features, $n$ is the number of samples and $d$ is the feature dimensionality. $\mathbf{A}$ is the ${n \times n}$ similarity matrix and $\mathbf{D}$ is the degree matrix which is a diagonal matrix with $D_{i, i}=\sum_{j} A_{i, j}$. $\mathbf{W}\in\R^{d\times d'}$ denotes the weight parameters of the convolution operator and $\mathbf{\widetilde{X}}$ is the output features.

When the parameter matrix $\mathbf{W}$ is an identity matrix, the graph convolution can be considered as feature propagation on the graph characterized by the similarity matrix. The mathematical formulation is
\begin{eqnarray} \label{eq:feature_update}
	\mathbf{\widetilde{X}}=\mathbf{D}^{-\frac{1}{2}}\mathbf{A}\mathbf{D}^{-\frac{1}{2}}\mathbf{X}
\end{eqnarray}
where $\mathbf{X}$ and $\mathbf{\widetilde{X}}$ have the same feature dimensionality. This operator is parameter-free, and can be directly used to update features based on existing feature vectors for visual retrieval tasks such as image retrieval and person Re-ID. 

For image retrieval, $\mathbf{X}$ contains image features from both the query set and gallery set. The GCN based operator explicitly propagates and updates features over a given graph. As a result, the retrieval results would be updated using the new features. It should be noted that most previous re-ranking methods directly update the ranked candidates or the original similarity matrix between the query set and the gallery set. In contrast, our approach implicitly renews the ranked results by updating and improving the features via a parameter-free graph convolution. 

As the GCN layers can be stacked to build deep architectures and learn high-level representations, the feature propagation can be iterated. Assume $T$ is the number of iterations. This process is represented by
\begin{eqnarray} \label{eq:iter_rank}
	\mathbf{X}_{t} = G(\mathbf{X}_{t-1})
\end{eqnarray}
where $G( \cdot )$ denotes the feature propagation in Eq.~(\ref{eq:feature_update}), $1\le t \le T$, $X_0$ is the original input features, and $\mathbf{X}_T$ is the final updated features.

The key of GCR is to design an appropriate similarity matrix. We impose locality constraints on the connected graph between the query and gallery samples. The details of feature propagation are elaborated in Section~\ref{feature_propgation}. 

\subsection{Feature Propagation on Graph} \label{feature_propgation}
Given the query and gallery samples, an undirected graph is constructed where each node denotes a sample and an edge denotes the relationship between two samples. The $k$-nearest neighbor graph is adopted to fully utilize the $k$-nearest neighbors of a sample for feature propagation. 

For an image sample index by $i$, we first obtain its $k$-nearest neighbors $\N_i$ based on the original features. Then, the similarity matrix $\widetilde{\mathbf{A}}$ is computed as	\begin{eqnarray}\label{similarity_matrix}
	\widetilde{A}_{i,j} = \left\{\begin{array}{lc}{\rm \exp}(-\|\mathbf{x}_i - \mathbf{x}_j\|_2^2/\gamma)&j\in\N_i\\1 & j=i\\0&\textrm{otherwise}\end{array}\right.
\end{eqnarray}
where $\mathbf{x}_i$ is the input feature of the $i\textrm{-th}$ image, and $\gamma$ is the temperature parameter. 

The similarity matrix $\widetilde{\mathbf{A}}$ originally obtained from Eq.~(\ref{similarity_matrix}) is not symmetric due to the fact that the $i\textrm{-th}$ image may not in the $k$-nearest neighbors $\N_j^k$ of the $j\textrm{-th}$ image when $j\in\N_i$. Two degree matrices can be computed by summing of the entries of the respective column or row. The formulations of row degree matrix $\mathbf{D_r}$ and column degree matrix $\mathbf{D_c}$ are
\begin{eqnarray}\left\{ \begin{array}{l}
		D_r(i,i) = \sum\nolimits_j {{\widetilde{A}_{i,j}}} \\
		D_c(j,j) = \sum\nolimits_i {{\widetilde{A}_{i,j}}} 
	\end{array} \right.
\end{eqnarray}

With the input features $\mathbf{X}$, the similarity matrix $\widetilde{\mathbf{A}}$, and degree matrices $\mathbf{D_r}$ and $\mathbf{D_c}$, the propagation criterion becomes
\begin{eqnarray}\label{eq:c1}
	\mathbf{\widetilde{X}}=\mathbf{D_{r}}^{-\frac{1}{2}} \mathbf{\widetilde{A}} \mathbf{D_{c}}^{-\frac{1}{2}} \mathbf{X}
\end{eqnarray}

As the similarity matrix contains pairwise similarity values which should be symmetric from an intuitive perspective. To make the similarity matrix symmetric, we simply add it with its transpose. The output symmetric matrix $A$ is
\begin{eqnarray} \label{eq:get_sym}
	\mathbf{A}=\left(\widetilde{\mathbf{A}}+\widetilde{\mathbf{A}}^{\top}\right) / 2
\end{eqnarray}

For the symmetric matrix, the row degree matrix and the column degree matrix are identical. Thus, the feature propagation criterion can be described in Eq.~(\ref{eq:feature_update}). The constructed nearest neighbor graph of the query and gallery samples is a sparse graph constrained on the nearest neighbors. 

The conventional feature propagation is computed with all images, which is inefficient to update features on a large-scale data set. To accommodate for large-scale retrieval, we present a decentralized and synchronous feature propagation method which is fit for parallel or distributed computing.
This method is summarized in Algorithm~\ref{alg:1}.

\begin{algorithm}[!ht]
	\caption{\textbf{D}ecentralized and \textbf{S}ynchronous \textbf{F}eature \textbf{P}ropagation (DSFP)}
	\label{alg:1}
	\textbf{Input:} $\mathbf{X}_0$ -- Original input features. \\
	\textbf{Output:} $\mathbf{X}_T$ -- Final updated features.\\
	\vspace{-12pt}
	\begin{algorithmic}[1]
	\STATE For the $i$-th image, obtain its $k$-nearest neighbors $\N_i$ with the provided features.
	\STATE Augment the $k$-nearest neighbors $\N_i$ with $\mathbf{x}_{i}, i.e.,  \widetilde{N}_i=N_i \cup \{i\}$.
	\STATE Generate the symmetric $(k+1)\times (k+1)$ local similarity matrix $\mathbf{A}^i$ for the $i$-th image as
	$$\forall u,v\in \widetilde{N}_i^k, A_{u,v}^i = \exp(-|\mathbf{x}_u - \mathbf{x}_v\|_2^2/\gamma)$$
	\vspace{-12pt}
	\STATE Prepare the local input features corresponding to $\widetilde{N}_i$, which is a $(k+1) \times d$ matrix denoted by $\mathbf{X}^i_0$.
	\STATE Calculate the local $(k+1)\times (k+1)$ degree matrix $\mathbf{D}^i$ with $D^i(i,i) = \sum\nolimits_j {A^i_{i,j}}$
	\STATE Iteratively update the features based on the feature propagation formulation,
		$$\mathbf{X}^i_{t+1}=(\mathbf{D}^i)^{-\frac{1}{2}}\mathbf{A}^i(\mathbf{D}^i)^{-\frac{1}{2}}\mathbf{X}^i_t$$
	where $1\le t \le T$, and $T$ is the number of iterations.
	\end{algorithmic}
\end{algorithm}

The proposed decentralized and synchronous feature propagation renew features for each image with a compact and local similarity matrix. 
Compared with the conventional global feature propagation, this distributed variant would be more efficient for large-scale applications.

\begin{figure*}
	\centering
	\includegraphics[width=0.85\textwidth]{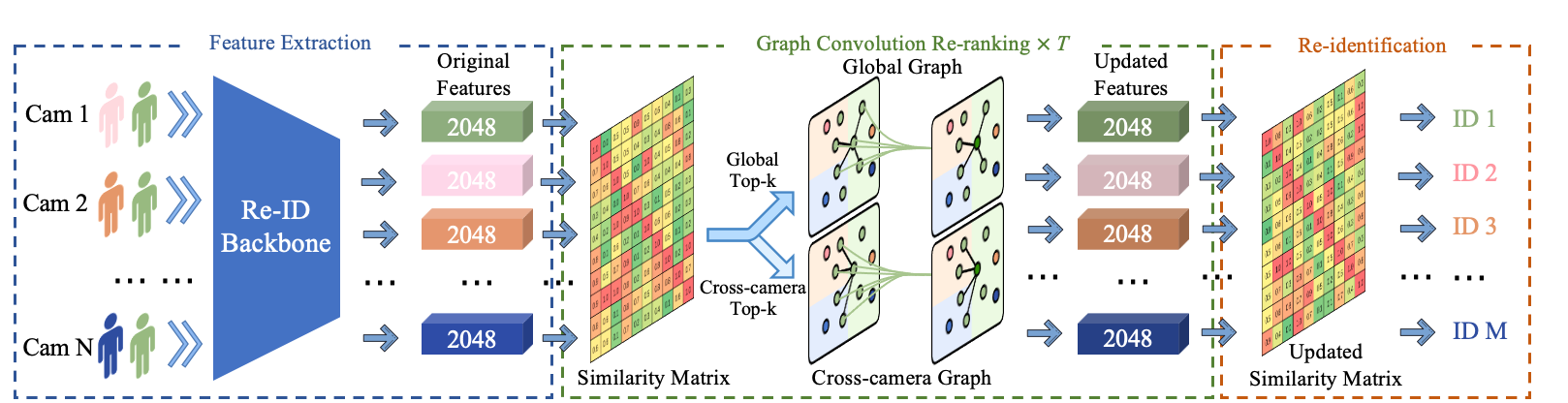}
	\caption{The pipeline of the proposed GCR applied for person re-identification.}
	\label{fig:framework}
\end{figure*}
\subsection{Extension for Cross-Camera Retrieval} \label{subsection:cross-camera}
For cross-camera retrieval tasks such as person Re-ID, samples from different cameras with similar contents should be emphasized while learning representations. 
We aim to align features from different cameras, which is very important for reducing feature divergences across different cameras. Cross-camera nearest neighbor graphs are additionally constructed only for samples from different cameras. The details are described as follows. 

For the $i$-th image, we first obtain its $k$-nearest neighbors $\N_i^{\textrm{cr}}$ from different cameras based on the original features.
We then compute the corresponding cross-camera similarity matrix $\widetilde{A}^\textrm{cr}$ as
\begin{eqnarray}
	\widetilde{A}^\textrm{cr}_{i,j} = \left\{\begin{array}{lc}\exp(-\|\mathbf{x}_i - \mathbf{x}_j\|_2^2/\gamma)&j\in\N_i^{\textrm{cr}}\\1&j=i\\0&\textrm{otherwise}\end{array}\right.
\end{eqnarray}
Similar to $\widetilde{A}$, the matrix $\widetilde{A}^\textrm{cr}$ is not symmetric, and the corresponding symmetric cross-camera similarity matrix $A^\textrm{cr}$ can be obtained in the similar way as Eq.~(\ref{eq:get_sym}).

With the two similarity matrices $\widetilde{\mathbf{A}}$ and $\mathbf{\widetilde{A}^\textrm{cr}}$, the feature propagation is
\begin{eqnarray}\label{eq:final}
	\widetilde{\mathbf{X}} = \alpha \mathbf{D_{r}}^{-\frac{1}{2}} \widetilde{\mathbf{A}} \mathbf{D_{c}}^{-\frac{1}{2}} \mathbf{X} + (1-\alpha)({\mathbf{D_{r}^\textrm{cr}}})^{-\frac{1}{2}} \mathbf{\widetilde{A}^\textrm{cr}} ({\mathbf{D_{c}^\textrm{cr}}})^{-\frac{1}{2}} \mathbf{X}
\end{eqnarray}
where $\alpha$ is the parameter to balance the weights between global propagation and cross-camera propagation. 
Similarly, the feature propagation formulation for symmetric similarity matrices $\mathbf{A}$ and $\mathbf{A^\textrm{cr}}$ can be deduced. 

The whole pipeline for the proposed GCR applied for person Re-ID is illustrated in Fig.~\ref{fig:framework}. As discussed in Eq.~(\ref{eq:iter_rank}), the re-ranking process can be iterated to get better representations. The similarity matrices as well as nearest neighbor graphs also change during different iterations. 


It should be noted that most existing methods~\cite{saquib2018pose,zhong2017re} leverage information from the global nearest neighbors to refine the results for person Re-ID, while building graphs only for nearest neighbors from different cameras is rarely investigated. Simple as it is, this step is important for cross-camera person Re-ID. Feature propagation on cross-camera graphs emphasizes the relationship between the image and its nearest neighbors from different cameras, and ultimately eliminates the bias from cameras in the similarity matrix and aligns features across multiple cameras.  

\subsection{Extension for Video-Based Retrieval} \label{subsection:video}
In subsection~\ref{subsection:cross-camera}, we extend the GCR to cross-camera retrieval scenarios. In this subsection, we further extend this approach to video-based retrieval, typically for video-based person Re-ID. 
Different from image-based re-identification, each sample in video-based re-identification is a tracklet, which consists of a set of images of a moving object. 

Suppose $\mathbf{x}_i$ and $y_i$ are the feature vector and identity label for the $i$-th image, respectively, and the tracklet which consists a sequence of images is indexed by the identity label.
Given images $\{\mathbf{x}_i, y_i\}$ from all tracklets in a video, a simple profile for the $c$-th tracklet is the mean feature vector
\begin{eqnarray}\label{eq:simple_profile}
	\hat{\mathbf{x}}_c = \frac{1}{m_c}\sum_{j|y_j=c} \mathbf{x}_j
	\label{eqn:meanvector}
\end{eqnarray}
where $y_j$ indicates the identification of the tracklet and $m_c$ is the number of images in the $c$-th tracklet. 

It is important to take full advantage of these multiple images in the tracklet to build a robust feature vector of this tracklet. Therefore, we propose a novel profile vector generation method to extract the profile vector for each tracklet. Based on the generated profile vector, the GCR can be easily applied for the video-based retrieval tasks.

We expect the new profile vector should be near to the features of images in the tracklet, and meanwhile far away from the other features in the same camera. Assume $\mathbf{w_c}$ is the profile vector of the ${c}$-th tracklet, a simple objective is to minimize the differences between $\mathbf{w_c}$ and images that share the same identity as the tracket and maximize the differences between $\mathbf{w_c}$ and other unrelated images. Thus, the objective function is formulated as
\begin{eqnarray}\label{eq:obj1}
	\min _{\mathbf{w_c}} \frac{1}{n m_c} \sum_{i} \sum_{j|y_{j}=c}\left\|\mathbf{x}_{j}-\mathbf{w_c}\right\|_{2}^{2}-\left\|\mathbf{x}_{i}-\mathbf{w_c}\right\|_{2}^{2}
\end{eqnarray}
where $m_{c}$ is the number of frames in the $c$-th tracklet, and $n$ is the total number of frames of images from all the tracklets. 

By rearranging terms and assume $\left\|\mathbf{x}_{i}\right\|_{2}=1$, the problem is equivalent to
\begin{eqnarray}
	\min _{\mathbf{w_c}} \frac{1}{n m_{c}} \sum_{i} \sum_{j|y_{j}=c} \mathbf{x}_{i}^{\top} \mathbf{w}_{c}-\mathbf{x}_{j}^{\top} \mathbf{w}_{c}
\end{eqnarray}

Inspired by the fact that the dot product between irrelevant vectors on the unit ball equals to 0, we focus on pushing $\mathbf{w}_{c}$ to its corresponding frames and reduce the influence from negative examples by employing a hinge loss
\begin{eqnarray}\min _{\mathbf{w_c}} \frac{1}{n m_{c}} \sum_{i} \sum_{j: y_{j}=c}\max(0, \mathbf{x}_{i}^{\top} \mathbf{w_c})-\mathbf{x}_{j}^{\top} \mathbf{w_c}\end{eqnarray}

By replacing the hinge loss with a smoothed version, a pair-wise smoothed hinge loss is defined as
\begin{eqnarray}
	\min _{\mathbf{w_c}} \frac{1}{n m_{c}} \sum_{i} \sum_{j|y_{j}=c}-\log \left(\frac{\exp \left(\mathbf{x}_{j}^{\top} \mathbf{w}_{c}\right)}{1+\exp \left(\mathbf{x}_{i}^{\top} \mathbf{w}_{c}\right)}\right)
\end{eqnarray}

According to Taylor expansion at $\mathbf{w}_{c}=0$, this minimized objective function can be bounded by the formulation
\begin{eqnarray}
	 \frac{1}{n} \sum_{i} \mathbf{x}_{i}^{\top} \mathbf{w}_{c}-\frac{1}{m_c} \sum_{j|y_{j}=c} \mathbf{x}_{j}^{\top} \mathbf{w}_{c}+ 
	\frac{1}{n}\left\|\mathbf{X} \mathbf{w}_{c}\right\|_{2}^{2}+\frac{\lambda}{2}\left\|\mathbf{w}_{c}\right\|_{2}^{2}
	\label{taylor_expansion}
\end{eqnarray}
where $\mathbf{X}$ is feature matrix which consists of all images from the tracklets, and $\left\|\mathbf{w}_{c}\right\|_{2}^{2}$ is a regularization term with a weight parameter $\lambda$.

The above quadratic objective function has a closed-form solution. Therefore, the formulation of the new profile is
\begin{eqnarray}\label{eq:profile}
	\mathbf{w}_{c}=\left(\mathbf{X} \mathbf{X}^{\top}+\lambda n \mathbf{I}\right)^{-1}\left(\frac{1}{m_{c}} \sum_{j| y_{j}=c} \mathbf{x}_{j}-\frac{1}{n} \sum_{i} \mathbf{x}_{i}\right)
\end{eqnarray}
where $\mathbf{I}$ is the identity matrix.


Compared with the mean vector in Eq.~(\ref{eqn:meanvector}), the profile vector in Eq.~(\ref{eq:profile}) eliminates the mean vector $\frac{1}{n}\sum_{i=1}^{n} \mathbf{x}_i$ of images of all tracklets from the same camera to reduce the bias from different cameras. It also leverages the geometric information by calculating the covariance matrix $\mathbf{X}^\top \mathbf{X}$.

After obtaining the profile vector, the proposed re-ranking method can be further used to re-rank video tracklets. The whole process is called Graph Convolution based Re-ranking for Video (GCRV), which is summarized in Algorithm~\ref{alg:2}.
\begin{algorithm}[!ht]
	\caption{\textbf{G}raph \textbf{C}onvolution based \textbf{R}e-ranking for \textbf{V}ideo (GCRV)}
	\label{alg:2}
	\textbf{Input:} A set of tracklets from videos of different cameras, $\mathbf{X}$ -- The corresponding original features  of images in the tracklets, $T$ -- The number of iterations.\\
	\textbf{Output:} $\mathbf{W}$ -- Updated profile vectors of all tracklets. \\
	\vspace{-12pt}
	\begin{algorithmic}[1]
		\STATE For each tracklet, generate the profile vector $\mathbf{w}_{c}$ as in Eq.~(\ref{eq:profile}).
		\FOR{$t=1$ {\bfseries to} $T$}
		\STATE For each tracklet, build two nearest neighbor graphs: global $k$-nearest neighbor graph and $k$-nearest neighbor graph from different cameras.
		\STATE Generate the global similarity matrices $\mathbf{A}$ and  cross-camera similarity matrix $\mathbf{\widetilde{A}^\textrm{cr}}$ based on profile vectors $\mathbf{W}=\{\mathbf{w}_{c}\}$.
		\STATE Update features $\mathbf{W}$ with the corresponding criterion as in Eq.~(\ref{eq:final}) by substituting $\mathbf{X}$ with $\mathbf{W}$.
		\ENDFOR
	\end{algorithmic}
\end{algorithm}

\section{Experiments}
\label{sec:exp}

\subsection{Datasets}
For image-based person ReID, we evaluate the proposed re-ranking approach on four popular benchmarks, i.e., Market-1501 (Market)~\cite{zheng2015scalable}, Duke-MTMC-re-ID (Duke)~\cite{ristani2016performance}, CUHK03~\cite{li2014deepreid} and MSMT17~\cite{wei2018person}.


For video-based person ReID, we evaluate the proposed approach on the MARS~\cite{zheng2016mars}, which is one of the largest dataset for video based person Re-ID. It consists of $17,503$ tracklets and $1,261$ identities. 

For image retrieval, we conduct experiments on the popular ROxford and RParis~\cite{radenovic2018revisiting} datasets. Each dataset contains 70 query images, which are divided into three splits: Easy, Medium and Hard. The Easy split ignores hard queries, the Medium split contains both easy and hard ones, while the Hard split only considers hard queries. 

\begin{table*}[!ht]
	\caption{Comparison with state-of-the-art methods for person Re-ID. }
	\label{tab:reid}
	\begin{center}
			\begin{tabular}{|l|c|cc|cc|cc|cc|}
				\hline
				\multirow{2}{*}{Method} & \multirow{2}{*}{Reference} &  \multicolumn{2}{c|}{Market}	& \multicolumn{2}{c|}{Duke}	 & \multicolumn{2}{c|}{MSMT17}	& \multicolumn{2}{c|}{CUHK03}\\    
				&	&  Rank-1 & mAP 	&  Rank-1 & mAP &  Rank-1 & mAP  &  Rank-1 & mAP \\
				\hline
				Deep GSRW~\cite{shen2018deep} & CVPR18 &  92.7 & 82.5 & 80.7 & 66.4 & -   &  -  & 94.9 & 94.0 \\
				SGGNN~\cite{shen2018person} & ECCV18 &  92.3 & 82.8  & 81.1 & 68.2 & - & - & 95.3 & 94.3 \\
				RED~\cite{bai2017ensemble}       		& CVPR19	& 94.7 & 91.0  	& -    & - 	 & -       & -      & -       & -  \\
				UED~\cite{bai2019re}       				& CVPR19	& 95.9 & 92.8  	& -    & - 	 & -       & -      & -       & -   \\
				SFT + KR~\cite{luo2019spectral}       	& ICCV19 	& 93.5 & 90.6   & 88.3 & 83.3  & 76.1    & 60.8   & 71.7    & 68.7 	\\
				SFT + LBR~\cite{luo2019spectral}       	& ICCV19 	& 94.1 & 87.5   & 90.0 & 79.6  & 79.0    & 58.3   & 74.3    & 71.7 \\
				ISP~\cite{zhu2020identity}              & ECCV20    & 95.3 & 88.6   & 89.6 & 80.0   & -       & -      & 76.5    & 74.1 \\
				MPN~\cite{ding2020multi}                & TPAMI20   & 96.3 & 89.4   & 91.5 & 82.0  & 83.5    & 62.7   & 85.0    & 81.1\\
				MGN + CircleLoss~\cite{sun2020circle}     & CVPR20    & 96.1 & 87.4   & -    & -      & 76.9    & 52.1   & -       & - \\
				Resnet50 + CircleLoss~\cite{sun2020circle} & CVPR20   & 94.2 & 84.9   & -    & -    & 76.3    & 50.2   & -       & - \\
				Deep-Person~\cite{bai2020deep} & PR20 & 92.3 & 79.6 & 80.9 & 64.8 \\
				OSNet~\cite{zhou2021learning} & TPAMI21 & 94.8 & 86.7 & 88.7 & 76.6 & 79.1    & 55.1   & 72.3    & 67.8 \\
				AGW~\cite{ye2021deep} & TPAMI21 & 95.1 & 87.8 & 89.0 & 79.6 & 68.3    & 49.1   & 63.6    & 62.0\\
				LAG-Net~\cite{gong2021lag} & TMM21 & 95.6 & 89.5 & 90.4 & 81.6 &  -   &  - & 85.1 & 82.2 \\
				MGN + FIDI~\cite{yan2021beyond} & TMM21 & 95.4 & 86.9 & 89.7 & 79.8 &  -  &  - & 78.9 & 76.3 \\
				OSNet + AutoLoss-GMS~\cite{gu2022autoloss} & CVPR22 & 95.7 & 88.9 & -- & --  & 83.7    & 62.6   & 74.5    & 72.6 \\
				RANGEv2~\cite{wu2022learning_reid} & PR22 & 94.7 & 86.8 & 87.0 & 78.2  & -       & -      & -       & - \\
				
				SDN~\cite{si2022spatial} & PR22 & 95.8 & 89.6 & 90.9 & 81.0  & -       & -      & 76.9    & 74.5 \\
				\hline
				BoT~\cite{luo2019bag} & CVPRW19   & 94.5 & 85.9 	& 86.5 & 76.4 & 63.4    & 45.1   & 58.0    & 56.6 \\
				BoT + KR~\cite{zhong2017re}         & -- 	& 95.4 & 94.2   & 90.2 & 89.1  & 64.3    & 57.7   & 68.7    & 68.1 \\
				BoT + ECN~\cite{saquib2018pose}     & -- & 95.8 & 93.2 	& 90.9 & 87.0  & 64.1    & 57.1   & 68.1    & 68.0	\\
				BoT + LBR~\cite{luo2019spectral} 	& -- & 95.7 & 91.3 	& 89.7 & 84.4 & 63.5    & 45.5   & 66.2    & 61.5\\
				BoT + GCR  & -- &  \textcolor{blue}{\textbf{96.1}}  &  \textcolor{blue}{\textbf{94.7}} & \textcolor{blue}{\textbf{91.5}} &  \textcolor{blue}{\textbf{89.7}}  & \textcolor{blue}{\textbf{64.9}}    & \textcolor{blue}{\textbf{58.5}}   &  \textcolor{blue}{\textbf{69.8}}    &  \textcolor{blue}{\textbf{68.4}} \\
				\hline
				MPN~\cite{ding2020multi} & TPAMI20    & 96.3 & 89.4   & 91.5 & 82.0 & 83.5    & 62.7   & 85.0    & 81.1  \\
				MPN + KR~\cite{zhong2017re}        & -- 	& 95.6 & 94.5   & 90.5 & 89.6  & 85.6    & 76.4   & 89.2    & 89.0 \\
				MPN + ECN~\cite{saquib2018pose}                    & --    & 95.1 & 94.0 	& 90.8 & 88.3 & 85.3    & 76.1   & 88.2    & 88.2 \\
				MPN + GCR	& -	& \textbf{96.6} &  \textbf{95.1} & \textbf{92.9} &  \textbf{91.3}  & \textbf{86.8}    & \textbf{76.9}   & \textbf{89.7}    & \textbf{89.0} \\
				\hline
			\end{tabular} 
	\end{center}
\end{table*}

\begin{table}[!ht]
	\caption{Comparison with state-of-the-art methods for video-based person Re-ID. }
	\label{tab:video_reid}
	\begin{center}
		\resizebox{0.72\linewidth}{!}{
			\begin{tabular}{|l|c|cc|}
				\hline
				\multirow{2}{*}{Method} & \multirow{2}{*}{Reference} &   \multicolumn{2}{c|}{MARS}       \\    
				&					&   Rank-1 & mAP \\
				\hline
				DSAN~\cite{wu2018and}  & TMM18 & 73.5 & -- \\ 
				AGRL~\cite{wu2020adaptive}              & TIP20    & 89.5 & 81.9 \\
				VKD~\cite{porrello2020robust}           & ECCV20    & 89.4 & 83.1 \\	
				MGH~\cite{yan2020learning}              & CVPR20    & 90.0 & 85.8 \\
				PhD~\cite{zhao2021phd}                   & CVPR21    & 88.9 & 86.2 \\
				STRF~\cite{aich2021spatio}                & ICCV21    & 90.3 & 86.1 \\
				DenseIL~\cite{he2021dense}              & ICCV21     & 90.8 & 87.0 \\
				PiT~\cite{zang2022multi}                   & TII22   & 90.2 & 86.8 \\
				SINet~\cite{bai2022salient} & CVPR22 & 91.0  & 86.2 \\ 
				MSTAT~\cite{tang2022multi} & TMM22 & 91.8 & 85.3 \\
				SGWCNN~\cite{yao2022sparse} & PR22 & 90.0 & 85.7 \\
				\hline
				BoT~\cite{luo2019bag}  & CVPRW19   & 85.8 & 79.7 \\
				BoT + KR~\cite{zhong2017re}         & --	& 84.3 & 85.2 \\
				BoT + ECN~\cite{saquib2018pose}     & -- & 88.1 & 85.6 \\
				BoT + LBR~\cite{luo2019spectral} 		& -- & 87.4 & 82.5 \\
				BoT + GCRV							&	--	  &  \textcolor{blue}{\textbf{89.0}} &  \textcolor{blue}{\textbf{87.0}}  \\
				\hline
				MGH~\cite{yan2020learning}				& CVPR20    &  90.0 & 85.8 \\
				MGH + KR~\cite{zhong2017re}                     & CVPR17 	& 88.8 & 90.7 \\
				MGH + ECN~\cite{saquib2018pose}           & CVPR18    & 92.7 & 90.5 \\
				MGH + GCRV 				    & -	&  $\textbf{93.8}$ &  $\textbf{92.8}$  \\
				\hline
			\end{tabular} 
		}
	\end{center}
\end{table}

\begin{table*}[t]
	\caption{Comparison with state-of-the-art methods for content-based image retrieval. }
	\label{tab:image_retrieval}
	\begin{center}
		\resizebox{0.68\linewidth}{!}{
			\begin{tabular}{|l|c|cc|cc|}
				\hline
				\multirow{2}{*}{Methods} &  \multirow{2}{*}{Reference} & \multicolumn{2}{c|}{Medium}   & \multicolumn{2}{c|}{Hard} \\
				& & ROxford 		& RParis 	 & ROxford	 &	RParis  \\
				
				\hline
				DSM~\cite{simeoni2019local}   & CVPR19 & 65.3 			& 77.4			& 39.2 			& 56.2\\
				IRT~\cite{el2021training} 	& \RED{arXiv21}	& 67.2 			& 80.1 			& 	42.8		& 60.5\\
				HOW+ASMK~\cite{tolias2020learning}  & ECCV20 & 79.4 			& 81.6			& 56.9			& 62.4\\
				DOLG~\cite{yang2021dolg}   & ICCV21  & 81.5 			& 91.0			& 61.1			& 80.3\\
				DALG~\cite{song2022dalg}     & \RED{arXiv22}   & 79.9 			& 90.0			& 57.6			& 79.1\\
				GRAP-CD~\cite{zhu2021graph}     & IVC21  & 70.8 			& 81.2			& 31.2			& 62.6\\
				MDA~\cite{wu2021learning}    & ICCV21   & 81.8 			& 83.3			& 62.2			& 66.2\\
				CSD~\cite{wu2022contextual}   & CVPR22  & 77.4 			& 87.9			& 59.0			& 75.7\\
				TBR~\cite{wu2022learning}   & AAAI22  & 82.3 			& 89.3			& 66.6			& 78.6\\
				\hline
				DELG global~\cite{cao2020unifying} & ECCV20 & 73.6 			& 85.7			& 51.0			& 71.5\\
				DELG global + $\alpha$QE~\cite{radenovic2018fine} & -- & 76.6 			& 86.7			& 54.6		& 73.2\\
				DELG global + GV~\cite{cao2020unifying}   & --   & 79.2 			& 85.5			& 57.5			& 67.2\\
				DELG global + RRT~\cite{tan2021instance}   & --   & 78.1 			& 86.7			& 60.2			& 75.1\\
				DELG global + GCR    & --    & \textcolor{blue}{\textbf{82.1}} 			& \textcolor{blue}{\textbf{89.2}}			& \textcolor{blue}{\textbf{63.1}}		& \textcolor{blue}{\textbf{72.4}} \\
				\hline
				Our global feature  & -- & 79.3 			& 84.4			& 62.8			& 74.4\\
				Our global feature + GCR       & --        & \textbf{84.3} 			&  \textbf{91.9}			&  \textbf{69.7}			&  \textbf{79.9} \\
				\hline
			\end{tabular}
		}
	\end{center}
\end{table*}

\subsection{Implementation Details}
For person Re-ID, the original features are extracted using the pre-trained models of ~\cite{luo2019bag, luo2019strong,liu2019spatially}, as \cite{luo2019bag} provides public pre-trained weights for the Market and Duke datasets, and \cite{liu2019spatially} provides pre-trained weights for the MARS dataset. The dimensionality of the extracted features is $2,048$. 

The default values of the temperature parameter $\gamma$, the number of nearest neighbors $k$, and the number of iterations $T$ are 0.2, 15 and 3, respectively. The weight parameter $\lambda$ of the regularization term in the profile vector generation is 10 on the MARS dataset.

\subsection{Results of Person Re-identification}
\label{sec:reid}
We first apply the proposed method for person Re-ID. The results on four popular datasets are summarized in Table~\ref{tab:reid}.

As representatives, we use the pretrained models of BoT~\cite{luo2019bag} and MPN~\cite{ding2020multi} on the corresponding dataset to extract features. GCR denotes our re-ranking method which utilizes both global and cross-camera neighborhood affinities for person Re-ID. 
To make a fair comparison, we also reproduce results of most commonly used re-ranking methods, i.e., k-reciprocal (KR)~\cite{zhong2017re}, ECN~\cite{saquib2018pose} and LBR~\cite{luo2019spectral}, based on the same features as those of our approach. 

It can be found that $BoT\!+\!GCR$ outperforms the origin baseline $BoT$ by a large margin. Meanwhile, $BoT\!+\!GCR$ has better performances those of other re-ranking methods for both Rank-1 and mAP and on four datasets.
We can also see that, augmented with state-of-the-art features such as MPN~\cite{ding2020multi}, our approach dramatically outperforms the recent state-of-the-art methods, and also consistently beats the other re-ranking methods. The results evidently demonstrate the superiority of our approach over previous re-ranking counterparts.

\subsection{Results of Video-Based Person Re-identification}
The results of video-based person Re-ID on the popular MARS dataset are provided in Table~\ref{tab:video_reid}. Recent state-of-the-art methods for this task are PIT~\cite{zang2022multi} and SINet~\cite{bai2022salient}, which utilize multi-direction and multi-scale pyramid in transformer and salient-to-broad transition with enhancement, respectively.
The proposed Graph Convolution based Re-ranking for Video-based retrieval is denoted as GCRV. 
We choose the recent representative video-based person Re-ID approach MGH~\cite{yan2020learning} and the widely used BoT~\cite{luo2019bag} as baselines.
Using BoT~\cite{luo2019bag} as the baseline, our approach significantly beats the widely used re-ranking approaches, i.e.,  KR~\cite{zhong2017re}, ECN~\cite{saquib2018pose} and LBR~\cite{luo2019spectral}, and improves the baseline by 3.2\% and 7.3\% for the criterion of Rank-1 and mAP, respectively. 
The mAP of our approach is even higher than that of the Transformer based PIT~\cite{zang2022multi} and the salient-to-broad transition based SINet~\cite{bai2022salient}.
While using the strong baseline MGH~\cite{yan2020learning}, our approach yields the state-of-the-art performance on this dataset and outperforms previous approaches by considerable margins.

\subsection{Results of Image Retrieval}
As the proposed plain GCR in Section~\ref{plain_gcr} is dedicated for general visual retrieval tasks, we conduct experiments on image retrieval to verify the effectiveness of this approach.
For content-based image retrieval (CBIR) such as building retrieval, a typical strategy is to first generate an initial rank list by global features. And then, local features are used to re-rank the candidates in the rank list. Due to the rigid characteristic of buildings, the re-ranking methods are often based on geometric verifications. In other words, two images should be semantically closer if they have more matched points generated by the affine operation. 
Typical re-ranking methods for CBIR include Query Expansion~\cite{radenovic2018fine} ($\alpha$QE) and geometric-based methods such as GV~\cite{cao2020unifying}, RRT~\cite{tan2021instance} and SuperGlue~\cite{sarlin2020superglue}. Different from person Re-ID, similarity or feature based re-ranking methods have seldom been exploited for building retrieval. 

The ROxford and RParis datasets focus on building retrieval where the buildings might be occluded or not salient in the image. This setting of CBIR is more complicated than person Re-ID, as images of Re-ID are tightly cropped. The results on the two datasets are given in Table~\ref{tab:image_retrieval}.
We first use the same global features as DELG~\cite{cao2020unifying}, and evaluate our approach as well as the re-ranking counterparts. 
As can be seen from the table, geometric-based re-ranking methods perform better than the naive similarity-based query expansion. 
However, the proposed GCR outperforms the geometric-based re-ranking. The results indicate that the internal appearance relationships among multiple building images are also important when compared with geometric relationships.

We also train a global image feature extractor based on~\cite{luo2019bag}. The proposed GCR based on these features achieves state-of-the-art performances on both datasets, outperforming the recent state-of-the-art approaches by large margins. Particularly, GCR improves the baseline of global features by 5.0\% and 7.5\% for the Medium split on the ROxford and RParis datasets, respectively. 

\begin{table}
	\caption{Effect of global or cross-camera affinity graphs in the proposed GCR for person Re-ID. For simplicity, Cross and Global denote the cross-camera similarity graph and global similarity graph, respectively. }
	\label{tab:ablation_graph}
	\begin{center}
		\resizebox{0.9\linewidth}{!}{
			\begin{tabular}{|l|cc|cc|}
				\hline
				\multirow{2}{*}{Method}	&\multicolumn{2}{c|}{Market}	&\multicolumn{2}{c|}{Duke}   \\ 
				&	Rank-1  & mAP		&       Rank-1  & mAP       \\
				\hline
				GCR w/o Cross    &    95.9    & 94.6     &       90.1    & 89.4  	\\
				GCR w/o Global	&    $\textbf{96.2}$    & 93.1    	&       $\textbf{91.6}$    & 86.8  \\
				GCR                &    96.0    & $\textbf{94.7}$     & 		91.5    & $\textbf{89.7}$	\\
				\hline
		\end{tabular}}
	\end{center}
\end{table}
\begin{figure}[t]
	\centering
	\begin{subfigure}[b]{0.475\linewidth}
		\centering 
		\includegraphics[width=\textwidth]{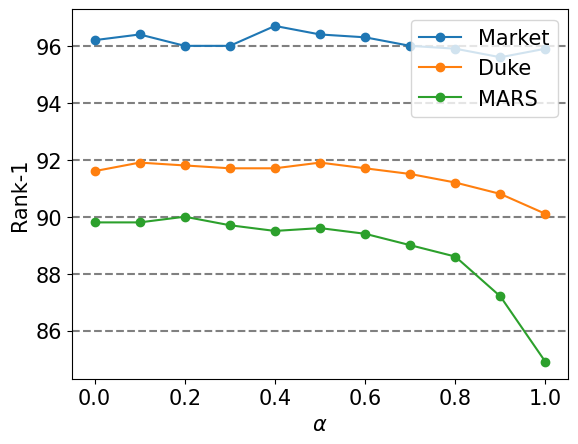}
		\caption{rank-1 curve}
	\end{subfigure}
	\begin{subfigure}[b]{0.475\linewidth}
		\centering                                                          
		\includegraphics[width=\textwidth]{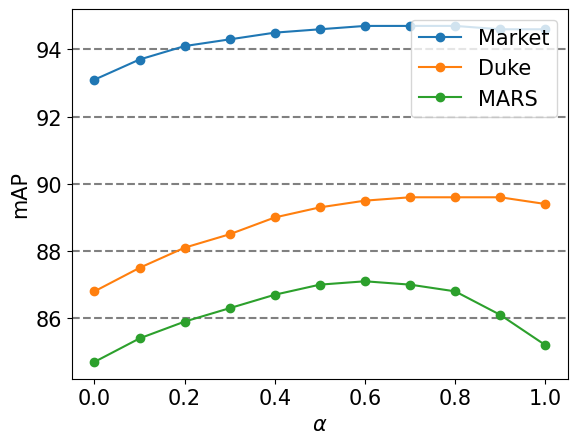}        
		\caption{mAP curve}
	\end{subfigure}
	\caption{The performance curves under different $\alpha$.}
\label{fig:alpha_tuning} 
\end{figure}

\begin{figure}
\centering
\includegraphics[width=0.5\textwidth]{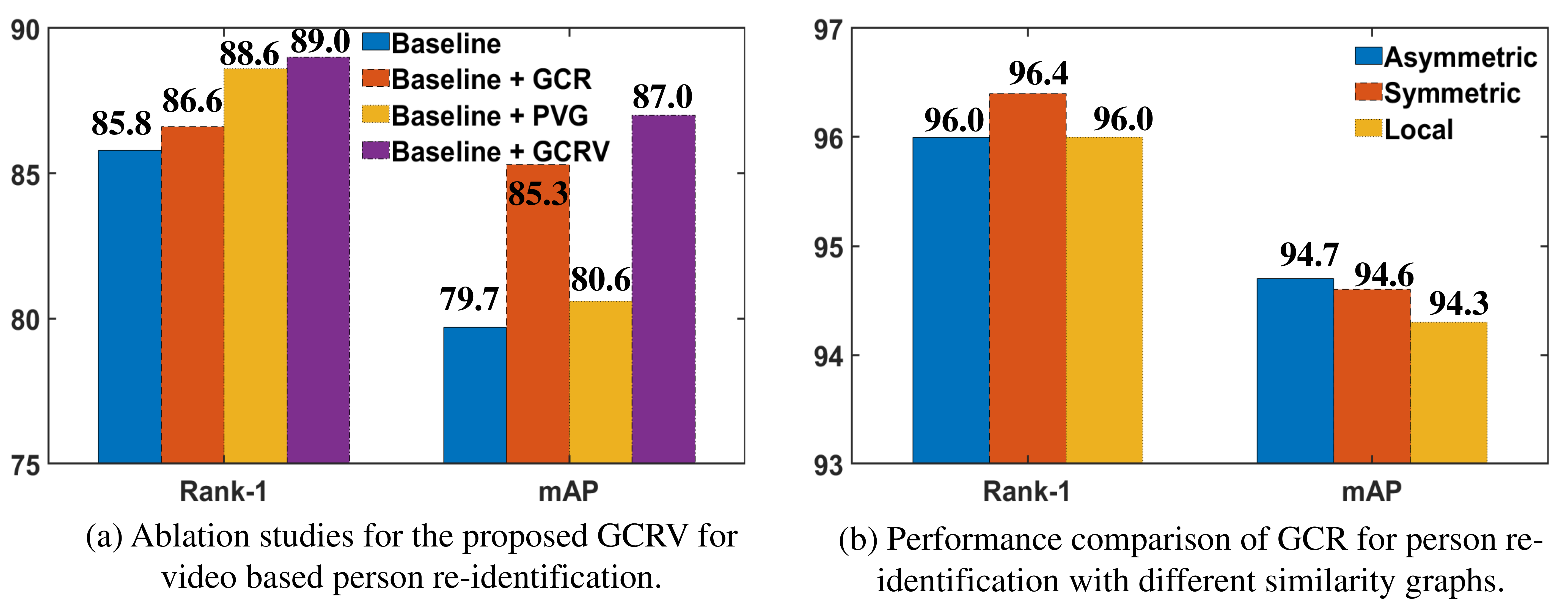}
\caption{Ablation studies for the GCRV and performance comparisons of the GCR with different graphs.}
\label{fig:ablation_vid}
\end{figure}
\begin{table}[t!]
	\centering
	\caption{The computation time of re-ranking methods on the Market-1501 dataset.}
	\begin{tabular}{|l|c|c|c|c|}
		\hline
		Method    &   KR~\cite{zhong2017re}  & ECN~\cite{saquib2018pose}  & GCR & GCR\_Local  \\
		\hline
		Time (s)   &   76         & 72   & 24  & 10     \\
		\hline
	\end{tabular}
	\label{tab:efficiency}
\end{table}

\begin{figure}[t]
\centering
\begin{subfigure}[b]{0.32\linewidth}
	\centering 
	\includegraphics[width=\textwidth]{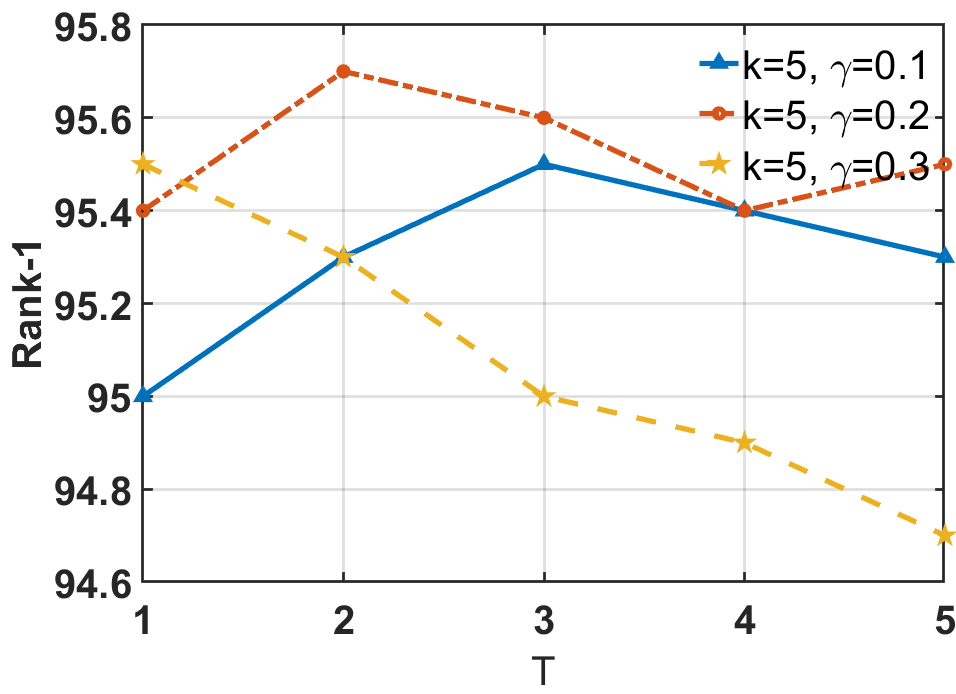}
	\caption{Rank-1 ($k=5$)}
\end{subfigure}
\begin{subfigure}[b]{0.32\linewidth}
	\centering                                                          
	\includegraphics[width=\textwidth]{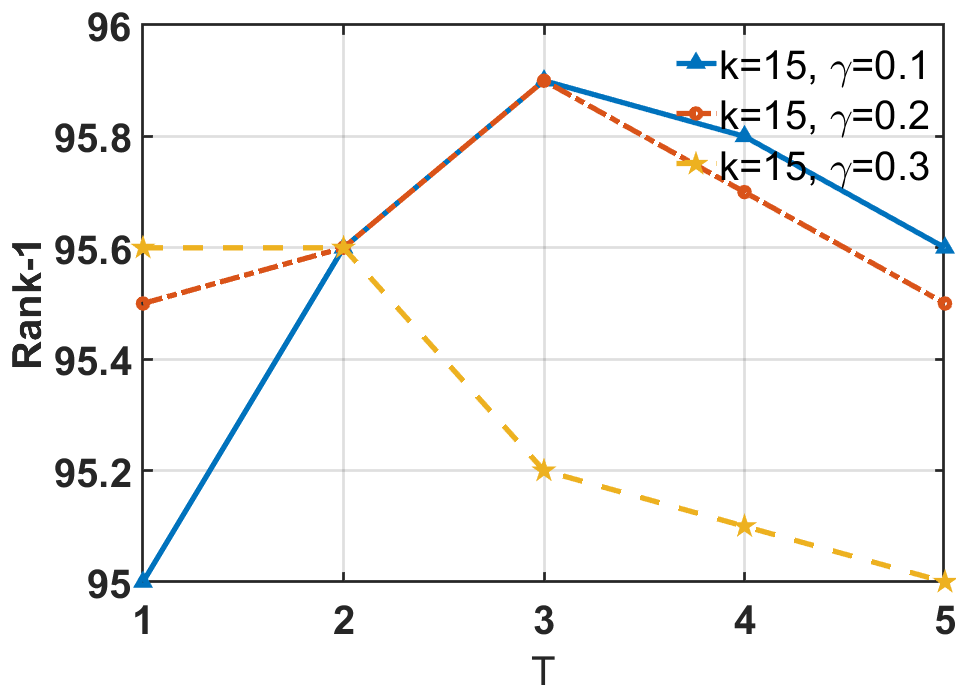}        
	\caption{Rank-1 ($k=15$)}
\end{subfigure}
\begin{subfigure}[b]{0.32\linewidth}
	\centering                                                          
	\includegraphics[width=\textwidth]{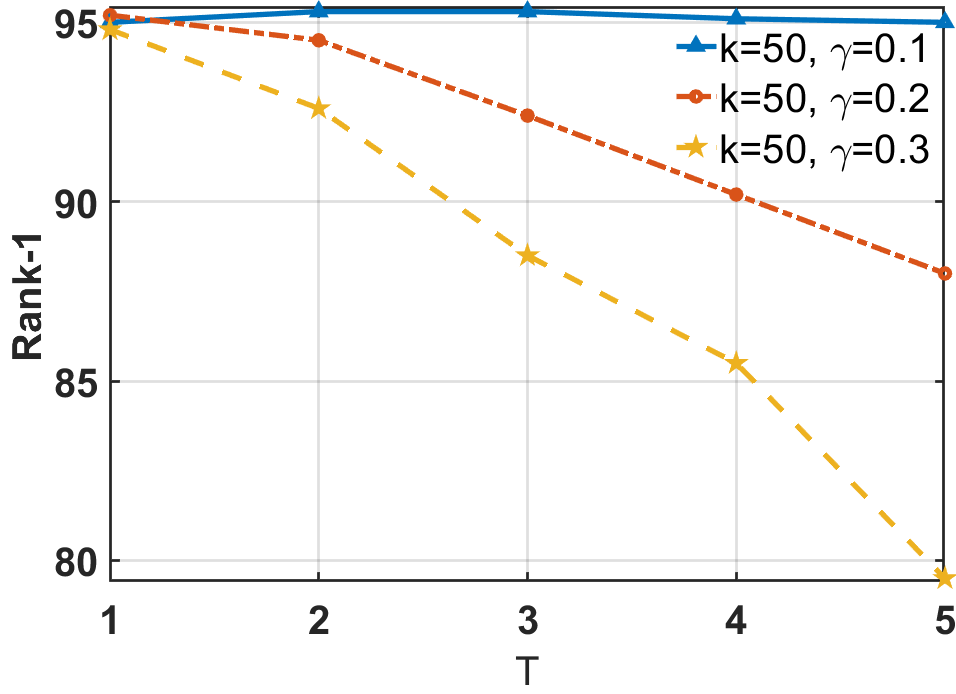}        
	\caption{Rank-1 ($k=50$)}
\end{subfigure}
\begin{subfigure}[b]{0.32\linewidth}
	\centering 
	\includegraphics[width=\textwidth]{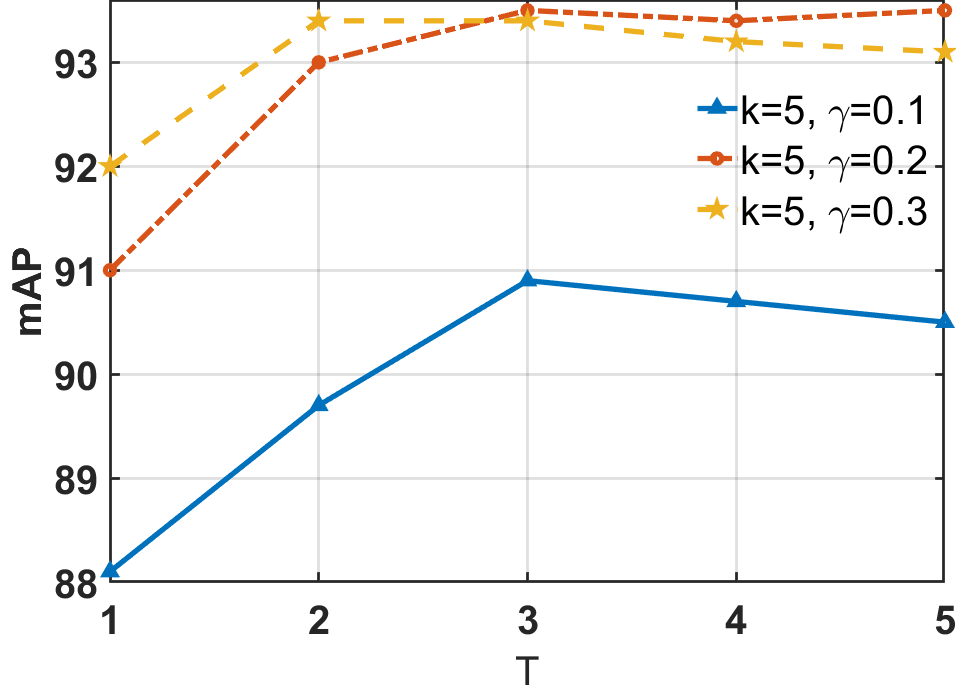}
	\caption{mAP ($k=5$)}
\end{subfigure}
\begin{subfigure}[b]{0.32\linewidth}
	\centering                                                          
	\includegraphics[width=\textwidth]{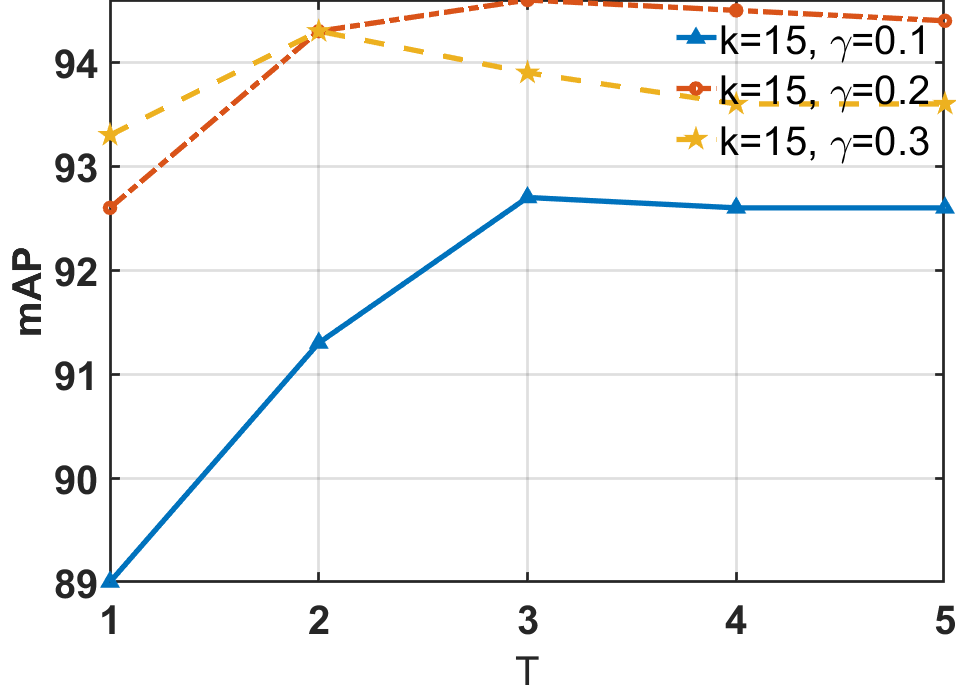}        
	\caption{mAP ($k=15$)}
\end{subfigure}
\begin{subfigure}[b]{0.32\linewidth}
	\centering                                                          
	\includegraphics[width=\textwidth]{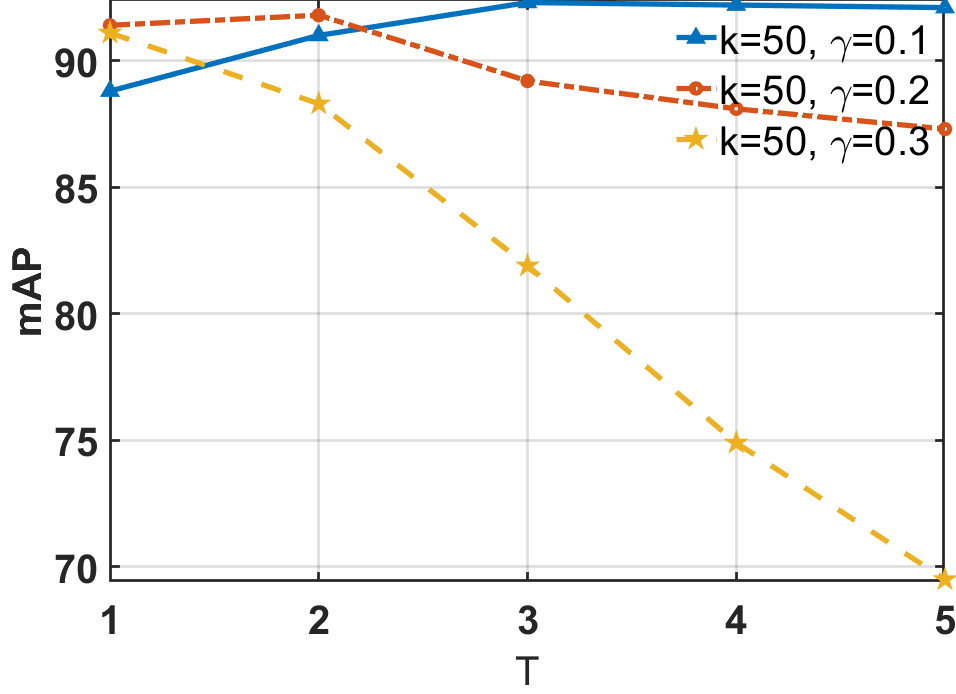}        
	\caption{mAP ($k=50$)}
\end{subfigure}
\caption{Sensitivity analysis of parameters in the proposed GCR.}
\label{fig:param_analy}
\end{figure}

\subsection{Experimental Analysis}
\subsubsection{Ablation Study}
We evaluate the effectiveness of the two nearest neighbor graphs for person Re-ID. The results are shown in Table~\ref{tab:ablation_graph}.
We find that only using the cross-camera similarity graph yields a high Rank-1, but a low value of mAP. In contrast, the global similarity graph achieves a high mAP. 
By combining the two graphs, ideal performance is achieved with both good performances of mAP and Rank-1 with the help of global and cross-camera similarity graphs, respectively. 

As shown in Eq.~(\ref{eq:final}), there is a trade-off hyper-parameter between the two graphs. We plot accuracy curves with respect to different $\alpha$ in Fig.~\ref{fig:alpha_tuning}. It can be seen that Rank-1 saturates for $\alpha < 0.7$, while mAP reaches the peak at $\alpha = 0.7$. Since mAP is often more important for retrieval cases, we select the hyper-parameter for the sake of better mAP and set $\alpha = 0.7$. 
As human pose and view orientation are two main factors that affect the visual appearance in the image, the global and cross-camera similarity graphs successfully align representations from both human poses and view orientations, thus getting the best retrieval performances.

The proposed GCRV for video based person Re-ID combines techniques of both Profile Vector Generation (PVG) and Graph
Convolution based Re-ranking (GCR). The ablated studies of GCRV on the MARS dataset are shown in Fig.~\ref{fig:ablation_vid}(a). The baseline uses the mean feature vector in Eq.~(\ref{eq:simple_profile}) to represent the tracklet, and does not adopt re-ranking. Baseline + GCR is the variant that uses the mean feature vector instead of the PVG. Baseline + PVG is the variant of GCRV without the GCR re-ranking. We find that PVG improves the performances of the mean feature vector for Rank-1 and mAP by 2.4\% and 1.7\%, respectively, which clearly demonstrates the effectiveness of the proposed profile vector generation and its superiority over the commonly used mean feature vector.

\begin{figure}[t]
	\centering
	\begin{subfigure}[b]{0.32\linewidth}
		\centering 
		\includegraphics[width=\textwidth]{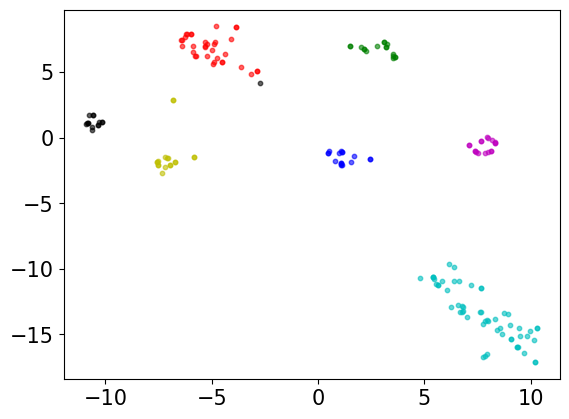}
		\caption{T=1}
	\end{subfigure}
	\begin{subfigure}[b]{0.32\linewidth}
		\centering                                                          
		\includegraphics[width=\textwidth]{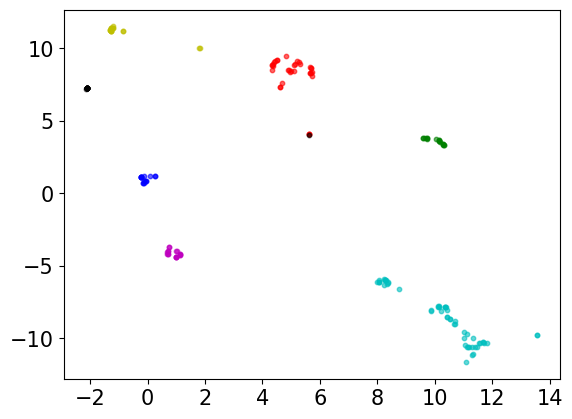}        
		\caption{T=3}
	\end{subfigure}
	\begin{subfigure}[b]{0.32\linewidth}
		\centering                                                          
		\includegraphics[width=\textwidth]{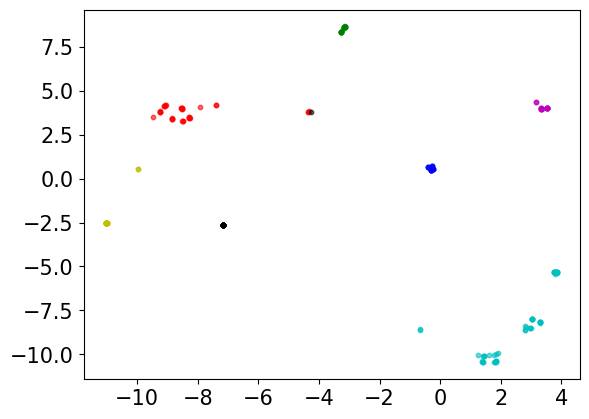}        
		\caption{T=5}
	\end{subfigure}
	\caption{Feature distributions for GCR with different iterations.\label{Fig.vis.main}}
\end{figure}
\begin{figure*}
	\centering
	\begin{subfigure}[b]{0.192\linewidth}
		\centering                                                          
		\includegraphics[width=\textwidth]{images/fac_round_3.png}        
		\caption{Ours}
	\end{subfigure}
	\begin{subfigure}[b]{0.192\linewidth}
		\centering                                                          
		\includegraphics[width=\textwidth]{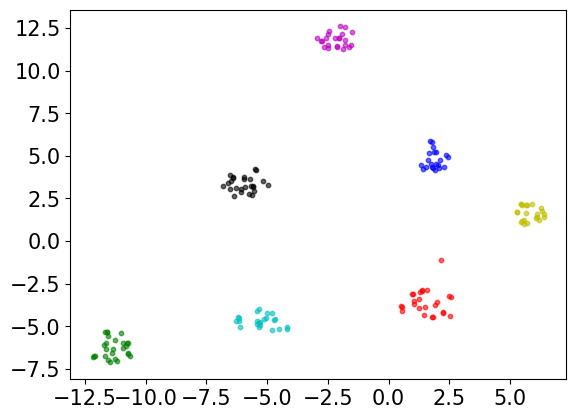}        
		\caption{KR~\cite{zhong2017re}}
	\end{subfigure}
	\begin{subfigure}[b]{0.192\linewidth}
		\centering                                                          
		\includegraphics[width=\textwidth]{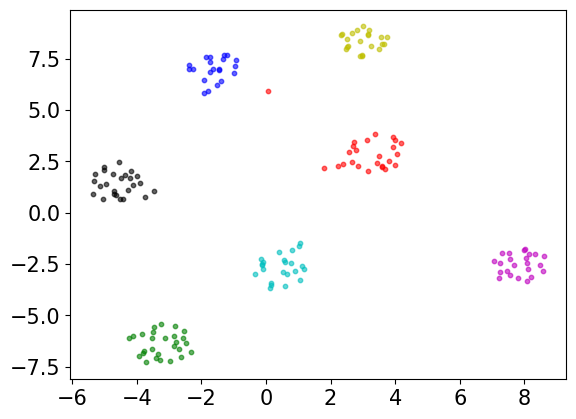}        
		\caption{ECN~\cite{saquib2018pose}}
	\end{subfigure}
	\begin{subfigure}[b]{0.192\linewidth}
		\centering                                                          
		\includegraphics[width=\textwidth]{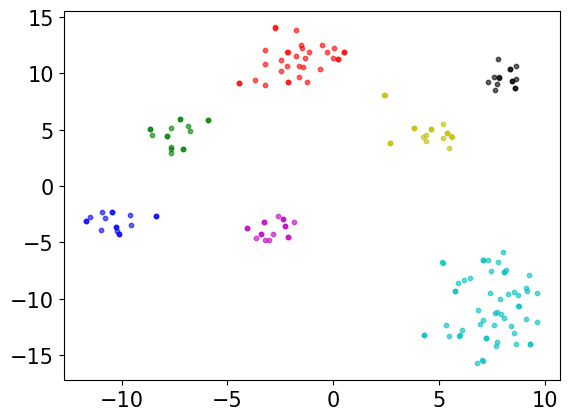}        
		\caption{LBR~\cite{luo2019spectral}}
	\end{subfigure}
	\begin{subfigure}[b]{0.192\linewidth}
		\centering                                                          
		\includegraphics[width=1.0\textwidth]{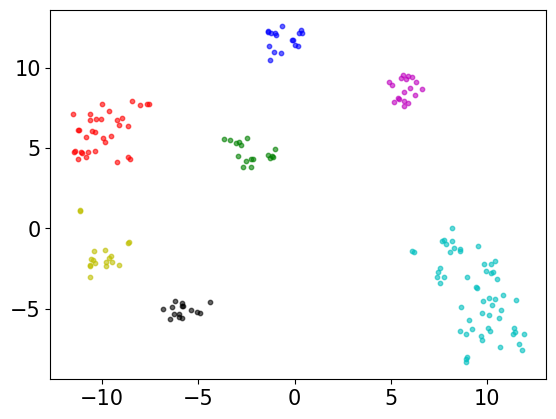}        
		\caption{QE~\cite{chum2007total}}
	\end{subfigure}
	\caption{Feature visualizations of different re-ranking methods.\label{Fig.vis_propogation}}
\end{figure*}

\begin{figure*} 
	\centering
	\includegraphics[width=1.0\textwidth]{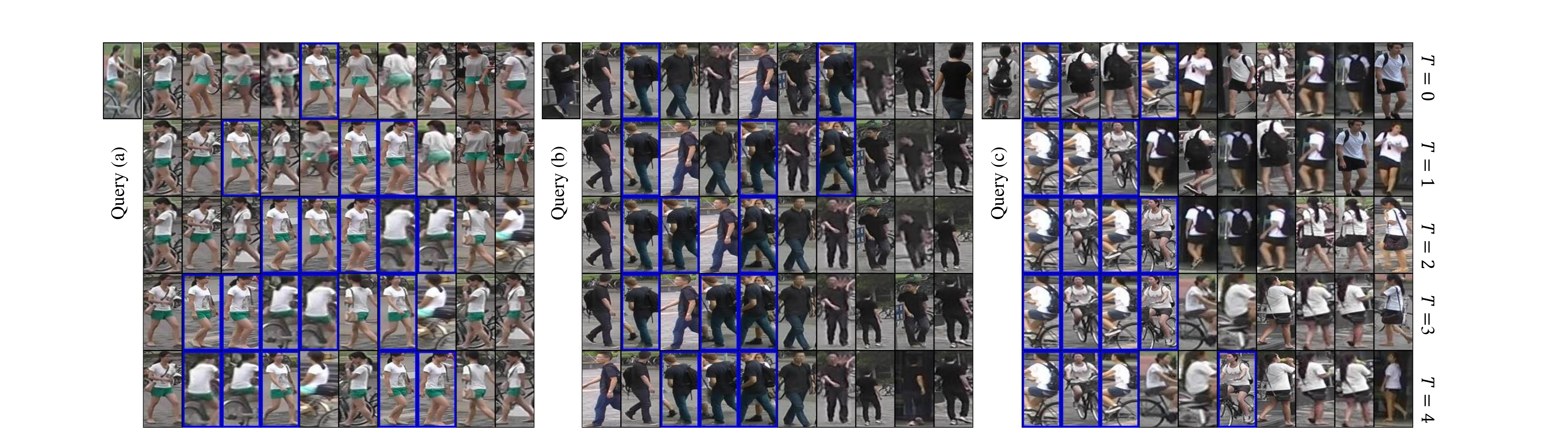}
	\caption{The visualization of results of person Re-ID for the proposed GCR with different iterations. $T = 0$ indicates the baseline without re-ranking, and correctly retrieved persons are marked with blue bounding boxes.}
	\label{fig:visualize_iter}
\end{figure*}    
\begin{figure*} 
	\centering
	\includegraphics[width=1.0\textwidth]{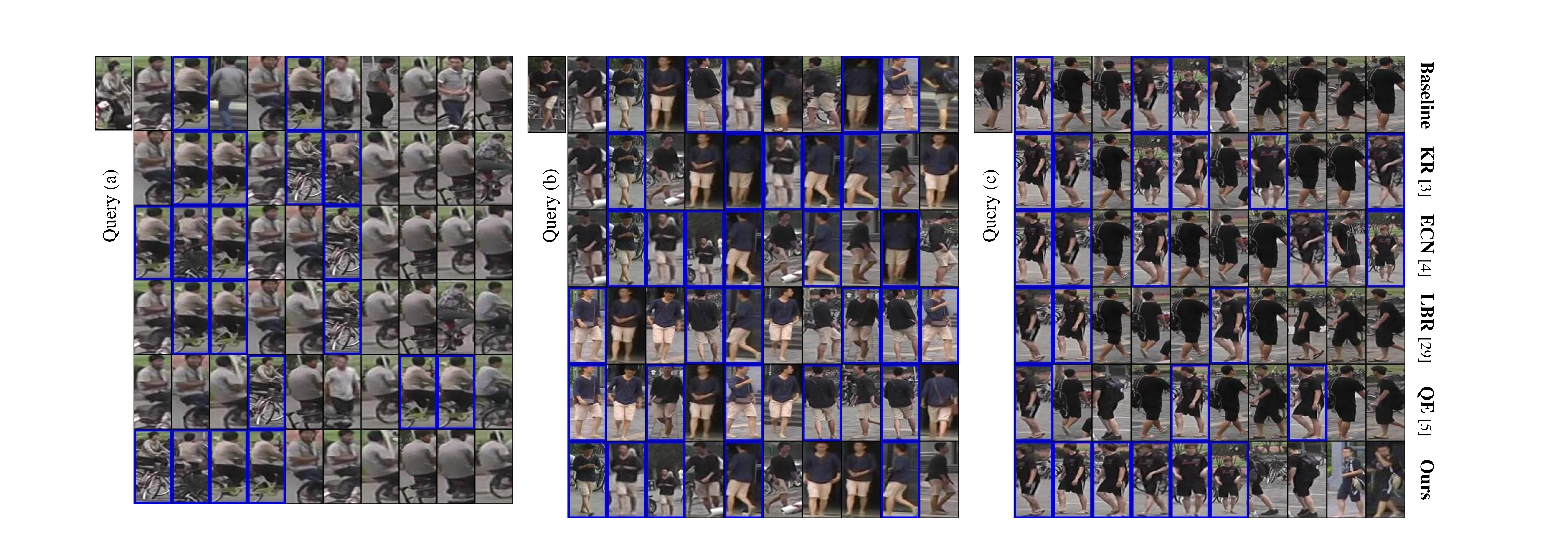}
	\caption{The visualization of results of person Re-ID for different re-ranking methods. The baseline involves no re-ranking. \RED{The top-10 retrieval results of the query are provided and correctly retrieved persons are marked with blue bounding boxes.}}
	\label{fig:visualize_stoa}
\end{figure*}  
\begin{figure*} 
	\centering
	\includegraphics[width=1.0\textwidth]{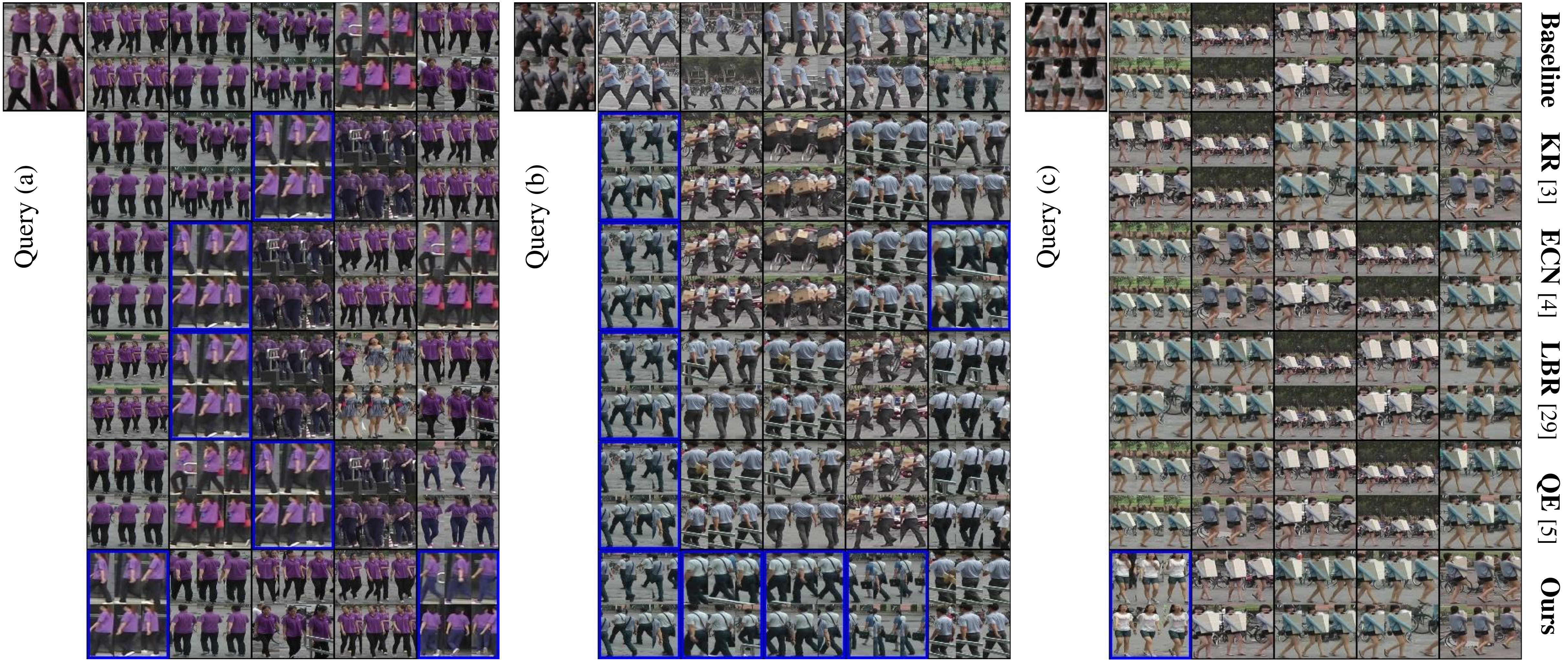}
	\caption{\RED{The visualization of results of video-based person Re-ID for different re-ranking methods on the MARS~\cite{zheng2016mars} dataset. Each video is represented by a sequence of six frames which are stitched into a large image. The top-5 results are provided from left to right and correctly retrieved videos of the query person are marked with blue bounding boxes.}}
	\label{fig:visualize_stoa_vid}
\end{figure*}  
\subsubsection{Efficiency Analysis} \label{efficiency}
In Algorithm~\ref{alg:1}, we propose a decentralized and synchronous feature propagation method based on local similarity matrices. This method is designed for parallel or distributed computing at the possible sacrifice of accuracy. Fig.~\ref{fig:ablation_vid}(b) shows performance comparisons of the proposed GCR for person Re-ID with different similarity graphs on the Market dataset. For simplicity, \emph{asymmetric}, \emph{symmetric} and \emph{local} denotes that the feature propagation adopts the asymmetric, symmetric and local similarity matrices, respectively. The performances of symmetric matrices are a little better than those of asymmetric ones. Compared with symmetric matrices, local matrices only slightly degrade the retrieval performance. The results demonstrate the effectiveness of the decentralized and synchronous feature propagation and suggest its vast potential applications for large-scale visual retrieval applications.

Efficiency is an important goal for algorithm design in real-world retrieval applications. Existing re-ranking methods such as KR~\cite{zhong2017re} suffer from the heavy computation. Table~\ref{tab:efficiency} lists the computation time of different re-ranking methods with the same hardware settings of 24 cores Platinum 8163 CPU. For simplicity, GCR\_Local denotes the decentralized and synchronous feature propagation variant. As can be seen, KR~\cite{saquib2018pose} and ECN~\cite{zhong2017re} suffer from low computation speed due to complex set operations. In contrast, our approach only relies on simple matrix operations and achieves both excellent accuracy and efficiency.

\RED{It should be noted that most of the computation costs of the proposed method focus on nearest neighbor graph construction which requires pairwise distance computing for all gallery pairs. Like other re-ranking methods such as KR~\cite{zhong2017re} and ECN~\cite{saquib2018pose}, the time complexity of the proposed GCR is $O({N^2}\log N)$, where $N$ denotes the number of samples in the database. The computation complexity of this variant in Algorithm~\ref{alg:1} is the same as the original method. However, nearest neighbor graphs in the database can be calculated and stored in advance, so the computation complexity can be reduced to $O(N \log N)$ in practical applications.
}

\subsubsection{Parameter Sensitivity}
There are three parameters in the proposed GCR: the temperature parameter $\gamma$, the number of nearest neighbors $k$, and the number of iterations $T$. The parameter sensitivity on the Market-1501 dataset is illustrated in Fig.~\ref{fig:param_analy}. We observe that a small value of $k$ leads to poor performance as the propagated information from such a small number of nearest neighbors is rather limited. In addition, the performance is saturated or even degraded when $k$ is too large (e.g., $k \geq 50$ ), as it increases the computational burden and introduces additional noises from semantically unrelated images. Thus, choosing an appropriate number of nearest neighbors is a bit important for the proposed method. For the sake of simplicity and good performance, we set $k = 15$ for all datasets. 

The best values of $\gamma$ and $T$ are somewhat related to the choice of $k$. Generally, when $k$ is small, a large $gamma$ and a relatively large $T$ (e.g., $T = 3$) is preferred to achieve good performance. And when $k$ is large (e.g., $k = 50$), smaller values of $\gamma$ and $T$ yield better performance. It can be interpreted that a smaller $\gamma$ lets the proposed method focus on more similar images, and hurts performance if only a small number of top nearest neighbors are selected. However, the setting of a smaller $\gamma$ would improve the performance when the nearest neighbors are redundant and noisy. 
More iterations allow feature propagation among more distant neighbors, which could increase the performance when $k$ is small and useful nearest neighbors are limited, however, decrease the performance when $k$ is large with the influence of noisy features.
Since $\gamma$ and $T$ are relatively not sensitive to the final performance, we keep them as $\gamma=0.2$ and $T=3$ in all experiments.

\RED{Changing the parameters such as the number of iterations does not affect the time complexity. Our approach accelerates the re-ranking process by avoiding inefficient computational operations and only using matrix operations. In addition, our approach is very suitable for GPU acceleration.}

\subsubsection{Visualization}
To further prove the superiority of the proposed method, we visually investigate the changes of features in the intermediate process of the GCR and compare visualized features among different feature propagation based re-ranking methods. Samples on the MARS dataset are selected and features are visualized using t-SNE~\cite{maaten2008visualizing}. 
Fig.~\ref{Fig.vis.main} shows the feature distributions for GCR with different iterations. 

It can be seen that features from the same person become more aggregated with the propagation of features. With more iterations $T>3$, the samples in blue are over-clustered into three points as seen in Fig.~\ref{Fig.vis.main}(c). Such over-clustering might cause matching mistakes. Similar results can be found in Fig.~\ref{fig:param_analy}, where more iterations $(T>3)$ no longer improve the accuracy.

Fig.~\ref{Fig.vis_propogation} visualizes \RED{and compares refined data points of different re-ranking approaches. }
Query Expansion (QE)~\cite{chum2007total} treats each neighbour equally, and LBR~\cite{luo2019spectral} performs local blurring on the gallery features while keeping query features unchanged. \RED{As can be seen, cyan data points in the bottom right corner of the Fig.~\ref{Fig.vis_propogation}(d) and \ref{Fig.vis_propogation}(e) are more scattered than those in Fig.~\ref{Fig.vis_propogation}(a).
It is indicated that}
both QE and LBR could not pull features of the same person similar enough to achieve discriminative representations. 
\RED{Data points of the same color in Fig.~\ref{Fig.vis_propogation}(a),~\ref{Fig.vis_propogation}(b) and~\ref{Fig.vis_propogation}(c) are well clustered and have compact distributions, which demonstrates the effectiveness of feature or distance updating. It should be noted that the input of the t-SNE is the updated similarity matrix instead of refined features for the re-ranking methods which do not involve feature propagation. }


We also compare the retrieval results of person Re-ID among different approaches with examples on the Market dataset. The results of the GCR with different iterations are shown in Fig.~\ref{fig:visualize_iter}. 
\RED{For both query (a) and (b), four of the top-5 results are correct when $T = 3$, while no more than three are correct for other counterparts. In general, }
the proposed re-ranking method pushes correct persons in the front of retrieved results, and $T = 3$ has the best overall retrieval results.

\RED{For person Re-ID}, retrieval results of different re-ranking methods are shown in Fig.~\ref{fig:visualize_stoa}. Compared with the widely adopted re-ranking methods such as KR~\cite{zhong2017re} and ECN~\cite{saquib2018pose}, our approach obtains comparable or even better refined ranked results.
\RED{In Fig.~\ref{fig:visualize_stoa}, the top-1 retrievals are all wrong except for ours for the query (a), while ours is the only approach where the top-3 retrievals are all correct for the query (b).}

\RED{Retrieval results of video-based person Re-ID are also shown in Fig.~\ref{fig:visualize_stoa_vid}. For the query (a) and (c), only our results are correct for the top-1 retrieval. Particularly, given the query (c) which contains images of the person taken from the back view, the other approaches fail while our approach correctly finds images of the same person taken from the front view. The results vividly demonstrate the effectiveness of our approach for visual retrieval tasks such as person Re-ID.
}

\section{Conclusion and Future Works}
\label{sec:conclude}
In this paper, we propose a general Graph Convolution based Re-ranking (GCR) method for visual retrieval. We extend this approach for cross-camera retrieval as well as video-based retrieval, and accordingly propose Graph Convolution based Re-ranking for Video (GCRV). Extensive experiments on three retrieval tasks, i.e., content-based image retrieval, person Re-ID, and video based person Re-ID evidently demonstrate the effectiveness of the proposed re-ranking method. Further ablation studies are analyzed to verify the effectiveness of the composed component, and parameter sensitivities are also conducted to analyze the robustness to hyper-parameters. 
Although this paper addresses the problems of inefficiency and inflexibility of existing re-ranking approaches, the performance of the variant for parallel computing deteriorates slightly. In the future, we will continue improving GCN based re-ranking and mainly focus on more efficient re-ranking algorithms on the GPU.

\ifCLASSOPTIONcaptionsoff
  \newpage
\fi

\bibliographystyle{IEEEtran}
\bibliography{egbib}

\end{document}